\documentclass[sn-basic]{sn-jnl}% Math and Physical Sciences Numbered Reference Style 
\usepackage{graphicx}%
\usepackage{multirow}%
\usepackage{amsmath,amssymb,amsfonts}%
\usepackage{amsthm}%
\usepackage{mathrsfs}%
\usepackage[title]{appendix}%
\usepackage{xcolor}%
\usepackage{textcomp}%
\usepackage{manyfoot}%
\usepackage{booktabs}%
\usepackage{algorithm}%
\usepackage{algorithmicx}%
\usepackage{algpseudocode}%
\usepackage{listings}%
\usepackage{comment}
\usepackage{array} 
\usepackage{varwidth} 
\usepackage{caption}
\usepackage{longtable}
\usepackage{paralist}
\usepackage{colortbl}
\usepackage{multirow}
\newcolumntype{P}[1]{>{\centering\arraybackslash}p{#1}}
\usepackage{enumitem}
\usepackage{booktabs}
\usepackage{csquotes}
\usepackage{hyperref}
 \usepackage[normalem]{ulem}
 \usepackage{soul}
 \usepackage{float}

\usepackage{pifont}
\newcommand{\cmark}{\ding{51}}
\newcommand{\xmark}{\ding{55}}

\begin{document}

\title[Article Title]{Exploring the Impact of Explainable AI and Cognitive
Capabilities on Users’ Decisions}


\author*[1]{\fnm{Federico Maria} \sur{Cau}}\email{federicom.cau@unica.it}

\author[1]{\fnm{Lucio Davide} \sur{Spano}}\email{davide.spano@unica.it}
\equalcont{These authors contributed equally to this work.}

\equalcont{These authors contributed equally to this work.}

\affil*[1]{\orgdiv{Department}, \orgname{Organization}, \orgaddress{\street{Via Ospedale 72}, \city{Cagliari}, \postcode{09124}, \state{Sardegna}, \country{Italia}}}


\abstract{

Artificial Intelligence (AI) systems are increasingly used for decision-making across domains, raising debates over the information and explanations they should provide. Most research on Explainable AI (XAI) has focused on feature-based explanations, with less attention on alternative styles.
Personality traits like the Need for Cognition (NFC) can also lead to different decision-making outcomes among low and high NFC individuals.
We investigated how presenting AI information (prediction, confidence, and accuracy) and different explanation styles (example-based, feature-based, rule-based, and counterfactual) affect accuracy, reliance on AI, and cognitive load in a loan application scenario. We also examined low and high NFC individuals' differences in prioritizing XAI interface elements (loan attributes, AI information, and explanations), accuracy, and cognitive load.
Our findings show that high AI confidence significantly increases reliance on AI while reducing cognitive load. Feature-based explanations did not enhance accuracy compared to other conditions. Although counterfactual explanations were less understandable, they enhanced overall accuracy, increasing reliance on AI and reducing cognitive load when AI predictions were correct.
Both low and high NFC individuals prioritized explanations after loan attributes, leaving AI information as the least important. However, we found no significant differences between low and high NFC groups in accuracy or cognitive load, raising questions about the role of personality traits in AI-assisted decision-making.
These findings highlight the need for user-centric personalization in XAI interfaces, incorporating diverse explanation styles 
and exploring multiple personality traits and other user characteristics to optimize human-AI collaboration.  

}




\maketitle

\section{Introduction}
\label{sec1}

Artificial Intelligence (AI) systems are becoming increasingly prevalent to assist human decision-makers across various domains, ranging from low-stakes activities like automating routine processes \citep{Herzog2019PlaceRec,Zehrung2021MovieRec,Musto2021FoodRec,Liao2022MoviesRec,Viswanathan2022PlaceRec,Kazjon2022FoodRec} to high-stakes scenarios like healthcare diagnostics \citep{Cai2019Medical,Min2020StrokeRehabilitationAssessment,Beede2020Medical,Min2021StrokeRehabilitationAssessment,Fogliato2022Medical,Panigutti2022Medical}. AI-assisted decision approaches pose numerous challenges within the HCI community, principally focusing on the problems of increasing users' accuracy and appropriate reliance on AI systems recommendations, i.e., accepting correct AI suggestions and rejecting wrong ones \citep{Zhang2020ConfidenceExplanationsAccuracyTrust, Rechkemmer2022AIPerformanceAndConfidenceEffectsOnUsers, Bove2022ContextualizationExplortationExampleBasedLocalFeatureImportance,Scharowski2023FeatureImportanceCounterfactualsExplanations,Kahr2023HumanLikeVsAbstractExplanations,Vasconcelos2023OverrelianceXAI,Chen2023RelianceExampleBasedFeatureBased}.  
In particular, previous research on human-AI teams mainly focused on investigating the following elements: task characteristics (e.g., complexity, stakes, and uncertainty) \citep{Bucinca2020ProxyTasks,Cau2023LogicalReasoningStock,Salimzadeh2023TaskComplexityDecisionMaking,Salimzadeh2024PrognosticVsDiagnosticTasks}, users' traits (e.g., Need for Cognition, task familiarity, and AI literacy) \citep{gajos2017NFC,Bucinca2021NFC,Gajos2022NFC, Ford2023ImageClassificationPeopleExpertiseExampleBasedExplanations,Celar2023CounterfactualCausal,He2023AnalogyExplanations}, the granularity of AI assistance (e.g., prediction, confidence, and accuracy) \citep{Yin2019AIAccuracyOnTrust,Lai2019ExampleBasedFeatureBasedAIPredictions,Zhang2020ConfidenceExplanationsAccuracyTrust,Rechkemmer2022AIPerformanceAndConfidenceEffectsOnUsers,Kahr2023HumanLikeVsAbstractExplanations,He2023AnalogyExplanations}, and explanation techniques to interpret AI decisions (e.g. example-based, feature-based, and counterfactuals) \citep{Lai2019ExampleBasedFeatureBasedAIPredictions,Bucinca2020ProxyTasks,Wang2022AIConfidenceMultipleExplanationsDomainExpertise,Bove2022ContextualizationExplortationExampleBasedLocalFeatureImportance,Chen2023RelianceExampleBasedFeatureBased,Teso2023InteractiveExplanations}. 
Despite these efforts, current research on AI-assisted decision-making exhibits diverging results on how and when AI assistance is delivered and which explanation styles could better help users assess the provided information. 

For example, presenting specific AI information (i.e., prediction, confidence, and accuracy) strongly influences users' decision-making processes. 
While showing predicted labels increases users' accuracy in the task than showing no AI assistance \citep{Lai2019ExampleBasedFeatureBasedAIPredictions,Bucinca2020ProxyTasks}, a high AI confidence (indicating the correctness likelihood in its prediction), appears to foster greater trust than a low one \citep{Zhang2020ConfidenceExplanationsAccuracyTrust,Rechkemmer2022AIPerformanceAndConfidenceEffectsOnUsers,Cau2023ExplImageText,Cau2023LogicalReasoningStock}. Additionally, a high stated AI accuracy on held-out data may affect people’s trust in the model by increasing their agreement with the AI \citep{Yin2019AIAccuracyOnTrust,Lai2019ExampleBasedFeatureBasedAIPredictions,Rechkemmer2022AIPerformanceAndConfidenceEffectsOnUsers,Kahr2023HumanLikeVsAbstractExplanations,He2023AnalogyExplanations,Kahr24AccuracyHumanExplanations}. Furthermore, studies on human-AI decision-making rarely evaluate users' cognitive load during task performance, and thus overlook the extent of cognitive resources being utilized \citep{SteyversKumar2024CognitiveEffort}.
The combined presentation of these AI information pieces and their influence on users' decision outcomes and perceptions is still understudied.

Another crucial aspect of the decision-making process involves eXplainable AI (XAI) techniques, whose potential to enhance user accuracy and appropriate reliance on AI is currently under debate.
While most empirical studies on AI decision support have focused on feature-based explanations \citep{Lai2023SurveyHumanAIDecisionMaking}, evidence remains inconclusive regarding their effectiveness in improving user accuracy or reducing overreliance \citep{Zhang2020ConfidenceExplanationsAccuracyTrust,Wang2021ExampleBasedFeatureBasedAndOthers,Shuai2023CorrectnessLikelihoodAIUsers,Cau2023LogicalReasoningStock,Chen2023RelianceExampleBasedFeatureBased}.
Additionally, while prior works have compared the effects of feature-based and example-based explanations on users \citep{Lai2019ExampleBasedFeatureBasedAIPredictions,CCAI2019ExampleBased,Bove2022ContextualizationExplortationExampleBasedLocalFeatureImportance,Ford2023ImageClassificationPeopleExpertiseExampleBasedExplanations,Chen2023RelianceExampleBasedFeatureBased,Lai2023SelectiveExplanations}, the benefits and limitations of other explanation styles, such as rule-based and counterfactual explanations, remain largely underexplored \citep{Wang2022AIConfidenceMultipleExplanationsDomainExpertise,Bodria2023Benchmarking,Teso2023InteractiveExplanations,Cau2023LogicalReasoningStock,Cau2023ExplImageText}.

Recent studies in music recommendation \citep{Millecamp2019NFC,Millecamp2020NFC}, AI-assisted nutrition decisions \citep{Bucinca2021NFC,Gajos2022NFC}, and intelligent tutoring systems \citep{Conati2021NFC,BahelConati2024NFC} have explored the influence of user-centric attributes like Need for Cognition (NFC) \citep{Cacioppo1984NFC} in user-AI teams. 
NFC is a personality trait that reflects an individual's tendency to engage in and enjoy effortful cognitive activities \citep{Carenini2001NFC,Cazan2014NFC,gajos2017NFC}. 
This research highlights significant differences in how low and high NFC individuals interact with AI, especially considering decision-making behavior, users' accuracy, reliance on AI, and cognitive load. 
While these studies provide some insights on specific domains, it is unclear how people with different NFC levels prioritize certain information in the XAI interface and how detailed AI information and multiple explanation styles affect their decisions.

Considering this, this paper investigates how including different AI information and explanations (i.e., prediction, confidence, accuracy, and explanation styles such as example-based, feature-based, rule-based, and counterfactual) impact users' decision-making process in a set of loan approval tasks considering their accuracy, reliance on AI, and cognitive load. Specifically, given the recent interest in studying the Need for Cognition (NFC) personality trait in human-AI teams, we aim to examine how different types of AI information
and explanation styles affect low and high NFC users in terms of i) how they prioritize the information in the XAI interface when making a decision, ii) the accuracy of the final decision, and iii) the required cognitive load.

\noindent
Our research questions to address these gaps are the following:

\begin{enumerate}[label = RQ\arabic*.]

    \item How do AI information and explanations impact users’ accuracy, reliance on AI, and cognitive load?
    
    \item Is there any difference in how people with low and high levels of Need for Cognition prioritize the information supplied in the XAI interface?

    \item Do people with low and high levels of Need for Cognition have different accuracy and cognitive load when engaging with explanations?

\end{enumerate}

To answer these questions, we conducted an online user study ($N$ = 288) where participants interacted with an AI-assisted loan approval interface, deciding whether to accept or reject eight loan requests based on varying AI assistance (i.e., no AI, AI with no explanation, AI with example-based, feature-based, rule-based, and counterfactual explanations). We analyzed their accuracy, reliance on  AI, cognitive load, and the importance of the XAI interface elements (i.e., loan attributes, AI information, and explanation) that led them to the final decision, further differentiating the results by low and high levels of Need for Cognition.

In summary, the contributions of this paper are:

\begin{enumerate}
\item We found that a high AI confidence significantly increases users' reliance on AI decisions while reducing cognitive load. These findings highlight the importance of calibrating AI confidence estimates to reflect the likelihood of system correctness. Additionally, integrating users' confidence calibration before AI interactions could enable new personalized AI-assisted strategies tailored to individual confidence levels.

\item Contrary to expectations, feature-based explanations did not improve user accuracy compared to other AI-assisted conditions. However, despite being perceived as less understandable by users, counterfactual explanations enhanced reliance on AI and reduced cognitive load, particularly when the AI predictions were correct, potentially improving overall accuracy. These findings suggest combining multiple explanation styles to complement each other's strengths and mitigate their shortcomings. 
This approach could lead to the development of personalized hybrid XAI visualizations.

\item We show that different levels (low and high) of personality traits like the Need for Cognition (NFC) might not capture differences in accuracy, cognitive load, and XAI interface element prioritization. 
While prior studies in less complex domains have often demonstrated differences in NFC levels, our results suggest that such distinctions may diminish as task complexity increases.
These findings suggest that NFC differences may not consistently generalize across diverse domains and tasks. Future studies should explore a broader range of personality traits and consider moving beyond personality-based factors to focus on other user-centric characteristics.
\end{enumerate}

Our paper is organized as follows.
We first review prior work on the influence of AI information, explainable AI (XAI) effectiveness, and the role of Need for Cognition (NFC) in AI-assisted decision-making (Sect. \ref{sec:related}). We then outline our hypotheses, further detailing the task design, including data, model, instances, and the AI assistance with explanations in Sect. \ref{sec:hypotheses}. We describe our study design, focusing on variables, sample size, statistical analysis, and the participants' procedure in Sect. \ref{sec:materials}. We present the results in Sect. \ref{sec:results}, beginning with descriptive statistics and hypothesis tests. This is followed by post hoc and exploratory analyses, covering task-specific metrics, interface understandability, and qualitative feedback. Next, we discuss the broader implications of our findings,  highlighting study limitations and proposing directions for future research in Sect. \ref{sec:discussion}. We conclude with key contributions and insights for improving personalized XAI systems in Sect. \ref{sec:conclusion}.
The study pipeline of data processing, model training, explanation generation, and statistical analysis is openly available at this link.

%% Related works
\section{Related Work}
\label{sec:related}
In this section, we first overview related work about the effectiveness of AI information and current eXplainable AI methodologies on users, considering the most common metrics to evaluate XAI systems and highlighting understudied topics. Then, we summarize previous studies on disaggregating low and high Need for Cognition participants in AI-assisted decisions focusing on the gaps of the current literature.

\subsection{Influence of AI Information on Decision Support}
\label{sec:related_ai}
Previous studies have shown that providing specific information about the AI assistant during decision-making (i.e., prediction, confidence score, and test set accuracy) strongly influences users' behaviors on task outcomes. 
For example, \citep{Lai2019ExampleBasedFeatureBasedAIPredictions} illustrated that showing predicted labels significantly improves
human performance in a deception detection task. They found that showing strong machine accuracy can induce similar human performance of featured-based explanations coupled with predicted labels.
Similarly, \citep{Bucinca2020ProxyTasks} found that participants who received AI predictions (with or without explanations) provided more accurate answers than those who did not receive any AI assistance in a nutrition-related decision-making task. 

As for AI confidence, \citep{Zhang2020ConfidenceExplanationsAccuracyTrust} explored its effects in an income prediction task and found that people trust the AI more in cases where the AI has higher confidence. Nevertheless, they found no evidence that AI confidence scores improve the accuracy of AI-assisted predictions. 
Another study from \citep{Rechkemmer2022AIPerformanceAndConfidenceEffectsOnUsers} showed that the effect of AI confidence on trust depends on people’s belief of the presented AI accuracy considering a speed dating event task. The higher the AI confidence, the more accurate people believe the model is. The authors argue that a possible reason for these results may lie in the users' perception of the AI information, considering AI accuracy as a fact and AI confidence as an estimate (i.e.,  less trustworthy than AI performance).
Additionally, \citep{Cau2023ExplImageText,Cau2023LogicalReasoningStock} found that low and high levels of AI confidence in predictions significantly affect users’ accuracy and agreement on AI, also influencing the effectiveness of different explanation styles considering different domains and stakes scenarios. 

Concerning the potential effects of AI accuracy on users, previous research \citep{Yin2019AIAccuracyOnTrust} explored how it affects people's trust in the model (i.e., agreement with the AI) in a speed dating task. The results show that high stated AI accuracy on held-out data increases people's trust in the model. Furthermore, trust is affected by both AI's stated accuracy and its observed accuracy during the task, and the effect of stated accuracy can change depending on the observed accuracy.  \citep{Rechkemmer2022AIPerformanceAndConfidenceEffectsOnUsers}  also found that AI's stated accuracy significantly increases people’s trust in the model in terms of agreement with the AI, switch fraction (i.e., users' change opinion after seeing the AI prediction), and self-reported trust in a second date prediction task. People trust the AI model more when its stated accuracy is higher. Additionally, the impact of the AI's confidence on people's belief in its predictions changes based on the AI's reported accuracy levels. 
A recent work from \citep{Kahr2023HumanLikeVsAbstractExplanations} also found that people's trust in the model is higher when presented with high-accuracy AI where users are asked to estimate jail time for 20 legal cases. 
In contrast, \citep{He2023AnalogyExplanations} found no significant effects of AI stated accuracy impacting users' reliance on the system  (expressed as agreement on  AI and switch fraction) in a loan prediction task.

To summarize, prior research consistently highlights that AI confidence and accuracy combinations affect users' reliance on AI during decision-making. We believe that when users are exposed to relatively high stated accuracy, the AI confidence acts as the tiebreaker in following the AI prediction: higher confidence increases the likelihood of users following the AI's suggestion.
Thus, this study explores the impact of AI information on user reliance on AI (i.e., agreement with AI decisions), particularly focusing on different levels of AI confidence. 
Furthermore, since users' cognitive load based on AI assistance is still underexplored in studies of AI-assisted decision-making \citep{SteyversKumar2024CognitiveEffort}, we argue that low AI confidence may elicit a higher cognitive load in users than high confidence, forcing them to reason independently rather than blindly following the AI's prediction.

\subsection{Explainable AI Effectiveness in AI-Assisted Decisions}
\label{sec:related_xai}
With the rise of complex black-box AI models, eXplainable AI techniques have emerged to help users understand how the AI reached a specific decision in low and high-stakes situations, including high-uncertainty and safety-critical contexts \citep{Bertrand2022BiasesInXAIMitigation,Lai2023SurveyHumanAIDecisionMaking,Rong2024HumanCenteredXAI,Harishankar2024XAIMedicalSurvey}.
Previous studies have shown that explanations may lead to increased user accuracy \citep{Lai2019ExampleBasedFeatureBasedAIPredictions,Bucinca2020ProxyTasks,Bansal2021FeatureBased,Herm2023impact} and appropriate reliance on AI \citep{Wang2022AIConfidenceMultipleExplanationsDomainExpertise,Scharowski2023FeatureImportanceCounterfactualsExplanations,Chen2023RelianceExampleBasedFeatureBased} when compared to AI prediction alone or not showing any assistance. 
Nevertheless, several studies on AI-assisted decisions explored explanation style differences in increasing users' accuracy and appropriate reliance, reporting contrasting results. 
Most of these studies focused on example-based and feature-based explanations \citep{Binns2018Loan,Lai2019ExampleBasedFeatureBasedAIPredictions,CCAI2019ExampleBased,Zhang2020ConfidenceExplanationsAccuracyTrust,Bove2022ContextualizationExplortationExampleBasedLocalFeatureImportance,Ford2023ImageClassificationPeopleExpertiseExampleBasedExplanations,Chen2023RelianceExampleBasedFeatureBased,Lai2023SelectiveExplanations}, with a limited number of studies also assessing the effects of rule-based and counterfactual explanations \citep{Gajos2022NFC,Wang2022AIConfidenceMultipleExplanationsDomainExpertise,Teso2023InteractiveExplanations,Celar2023CounterfactualCausal}. 
For example,
\citep{Wang2022AIConfidenceMultipleExplanationsDomainExpertise} studied the effects of different explanations (i.e., feature importance, feature contribution, nearest neighbors, and counterfactuals) in a recidivism prediction task, 
and found that when users have
some domain expertise in the decision-making task, feature contribution can satisfy more desiderata of AI model and explanations (i.e., understanding, uncertainty awareness, and trust calibration)
regardless of the complexity of the AI model.
Another study from \citep{Chen2023RelianceExampleBasedFeatureBased} found that example-based explanations for an income prediction task increased accuracy with AI correct predictions than showing no AI assistance. Instead, when the AI was incorrect, the authors found a trend of feature-based explanations increasing overreliance.
Furthermore, \citep{Cau2023LogicalReasoningStock} investigated the effects on AI confidence and logic-style explanations in a stock trading market task, discovering that when AI confidence is high, users tend to over-rely on an erroneous AI more with inductive (example-based) explanations than abductive (feature-based) and deductive (rule-based) explanations.

Given that most of the existing XAI literature has focused on feature-based explanations \citep{Lai2023SurveyHumanAIDecisionMaking}, and there is insufficient evidence regarding their impact on users' accuracy, particularly with tabular data \citep{Zhang2020ConfidenceExplanationsAccuracyTrust,Wang2021ExampleBasedFeatureBasedAndOthers,Chen2023RelianceExampleBasedFeatureBased,Shuai2023CorrectnessLikelihoodAIUsers,Cau2023LogicalReasoningStock}, we aim to investigate whether feature-based explanations improve users' accuracy compared to other types of AI assistance (i.e., no AI, AI, example-based, rule-based, and counterfactual explanations).

\subsection{Need for Cognition in Human-AI Decisions}
\label{sec:related_nfc}
Need for Cognition (NFC) \citep{Cacioppo1984NFC} is a measure that reflects the tendency for an individual to undertake effortful cognitive activities \citep{gajos2017NFC,Bucinca2021NFC} and benefit more from complex user interface features \citep{Carenini2001NFC,Cazan2014NFC, gajos2017NFC, Ghai2021NFC,Gajos2022NFC}.
Previous work has shown that people with higher NFC are more likely to be curious and in a focused attentive state while using a computer \citep{Li2006NFC}, and have higher performance at complex skill acquisition in the context of computer task performance \citep{Day2007NFC}.
Considering explanations in music recommendations (i.e., assisted creation of a playlist), \citep{Millecamp2019NFC} found that explanations raised users' confidence with a low NFC when making their playlist. In contrast, users with a high NFC experienced a decrease in their confidence due to explanations. On the contrary, a follow-up study from \citep{Millecamp2020NFC} did not find an effect of NFC on the perception of explanations. The authors stated that a potential reason for this result might lie in the explanations presentation and the proactive activation of explanations which brings out the differences between low and high NFC users. While in the previous study \citep{Millecamp2019NFC} explanations had to be explicitly activated by the users, in \citep{Millecamp2020NFC} explanations were always visible. 

Concerning NFC effects in the nutrition domain, \citep{Bucinca2021NFC} studied the impact of cognitive forcing functions (i.e., interventions that disrupt heuristic reasoning and cause the person to engage in analytical thinking) and simple XAI approaches among low and high NFC participants in an AI-assisted nutrition study (e.g., making a plate low-carb by changing the ingredients accompanied by AI and explanations) with a simulated AI. 
Despite high NFC participants trusting and preferring cognitive forcing functions less than simple explainable AI approaches, they generally performed better in the task than low NFC participants.
Furthermore, low NFC participants generally found the task significantly more mentally demanding and the system considerably more complex than high NFC participants.
This might confirm the findings from \citep{Millecamp2019NFC,Millecamp2020NFC} that only cognitive forcing functions produce intervention-generated inequalities between people based on their NFC level. 

Another study on AI-assisted nutrition by \citep{Gajos2022NFC} found that explanation-only design (without AI recommendation and before the user decision) benefits people with a high NFC more in task learning than those with low NFC.  
This finding contrasts with previous studies, suggesting that differences in participants with diverse levels of NFC may emerge without using interventions like cognitive forcing functions. 
In the context of AI-assisted maze solving, a recent study from \citep{Vasconcelos2023OverrelianceXAI} investigated whether overreliance was affected by the interaction between participants’ NFC scores and the AI with and without explanations when the task was hard to solve (both the AI and explanations were simulated). However, they did not find any evidence, potentially because the hard task was too difficult to demonstrate differences in behavior across participants' NFC since most people are likely to over-rely on the AI’s prediction anyway.

Based on this body of research, our work aims to deepen the alleged requirement for cognitive forcing functions to highlight the differences between low and high NFC participants. Specifically, apart from \citep{Gajos2022NFC} results, the use of interventions to provide explanations to users on-demand or employing two-stage detection paradigms \citep{Green2019TwoStageDetection,Green2019TwoStageDetection2,He2023AnalogyExplanations} where users make the initial decision alone and then make a second final choice to decide whether to incorporate AI advice seem to be the only ways to elicit differences in low and high NFC participants. Additionally, previous studies investigating participants' NFC
used simulated AIs, always correct AI's recommendations, and one/two types of simulated explanations.
Therefore, we examine whether a difference exists between low and high NFC participants' decision-making given different AI information and explanations (i.e., prediction, confidence, accuracy, and explanation styles such as example-based, feature-based, rule-based, and counterfactual) in a complex \citep{Salimzadeh2023TaskComplexityDecisionMaking} and high-stakes loan application scenario considering users' accuracy, cognitive load, and how they prioritize the XAI interface information.

\section{Hypotheses and Task Design}
\label{sec:hypotheses}

In this section, we start describing how we translated our research questions into hypotheses, studying how AI information and explanations affect decision-making (RQ1), how individuals with varying levels of Need for Cognition prioritize interface elements (RQ2), and whether these individuals differ in accuracy and cognitive load (RQ3). We then detail the task design scenario employed to test these hypotheses.

\subsection{Hypotheses}
\noindent\textit{Hypotheses Related to RQ1.} 
As discussed in Section \ref{sec:related_ai}, previous research indicates that low and high levels of AI confidence and accuracy affect user reliance on AI in decision-making. Given we showed users a fixed AI accuracy that is relatively high (i.e., 83\% on the test set, see Section \ref{sec:model_}), we believe that high AI confidence will lead users to rely more on AI predictions. Conversely, low AI confidence may encourage users to think independently, increasing their cognitive load compared to high AI confidence. 
In Section \ref{sec:related_xai}, we also mentioned that previous work does not highlight any strong advantages of rule-based and counterfactual explanations over feature-based ones. Additionally, the efficacy of example-based explanations primarily depends on the similar instances retrieved. Given that we are considering tabular data, presenting similar instances would significantly increase task complexity and thus users' cognitive load \citep{Salimzadeh2023TaskComplexityDecisionMaking,Cau2023LogicalReasoningStock}, which may lead them to rely on the most frequent AI prediction across the similar instances (such as accepting if the majority of similar instances are accepted) rather than carefully analyzing each instance individually.
Instead, feature-based explanations (in our case, feature contribution) provide users with an immediate overview of important attributes relevant to the AI's decision and seem at a glance to satisfy more desiderata for AI models and explanations (i.e., understanding, uncertainty awareness, and trust calibration) when users people are somewhat knowledgeable about the target domain \citep{Wang2022AIConfidenceMultipleExplanationsDomainExpertise}.
Although satisfying more desiderata does not imply an increased accuracy in the task, we hypothesize that feature-based explanations might lead users to achieve higher accuracy than the other AI assistance conditions.
Summarizing, we formulate the following hypotheses:

\begin{itemize} %

    \item \textbf{H1a}: Users exposed to a high AI confidence will rely more on the AI prediction than users exposed to a low AI confidence.
    
    \item \textbf{H1b}: Users exposed to a high AI confidence will report a lower cognitive load than users exposed to a low AI confidence.

    \item \textbf{H1c}:  Users exposed to feature-based explanations will achieve higher accuracy than other AI assistance conditions.

\end{itemize}

\noindent\textit{Hypotheses Related to RQ2.}
\noindent As mentioned in Section \ref{sec:related_nfc} and given the high complexity of the loan prediction task coupled with AI explanations, we hypothesize that low NFC participants might base their decisions mostly on AI information rather than interpreting the explanations to complete the task. Instead, given the tendency of high NFC participants to
enjoy cognitively demanding activities, they will try to inspect the explanation to gather additional insights about the validity of the AI information to complete the task. Hence, we formalized the following hypotheses:

\begin{itemize}

    \item \textbf{H2a}: Users with a low NFC will mainly prioritize the applicant's details to make their final decision (rank 1), then the AI information (rank 2), and lastly the explanation (rank 3). 
    
    \item \textbf{H2b}:
    Users with a high NFC will mainly prioritize the applicant's details to make their final decision (rank 1), then the explanation (rank 2), and lastly the AI information (rank 3). 

\end{itemize}

\noindent\textit{Hypotheses Related to RQ3.}
\noindent We hypothesize that high NFC participants will leverage explanations to get more insights about the information provided by the AI, potentially achieving higher accuracy than the low NFC ones. Additionally, given their inclination to enjoy complex cognitive activities, high NFC participants will report a lower cognitive load in completing the loan approval tasks:

\begin{itemize} 
    \item \textbf{H3a}:
    When provided with explanations, users with a high NFC will achieve a higher accuracy than users with a low NFC. 

    \item \textbf{H3b}: When provided with explanations, users with a high NFC will report a lower cognitive load than users with a low NFC.

\end{itemize}

\subsection{Task Design}
This subsection defines how we implemented the loan application task, describing the data we used, the model, instances selection, and model explanation generation.

\subsubsection{\textbf{Data}} 
We built the loan approval task on the publicly available \textit{Loan Prediction Problem Dataset}\footnote{\href{https://www.kaggle.com/datasets/altruistdelhite04/loan-prediction-problem-dataset}{https://www.kaggle.com/datasets/altruistdelhite04/loan-prediction-problem-dataset}}, consisting of 614 loan requests where the goal is to decide whether to accept or reject a loan application based on twelve features.
We opted for this dataset since it reflects a realistic and fairly complex human-AI collaboration scenario \citep{Salimzadeh2023TaskComplexityDecisionMaking,He2023AnalogyExplanations}. Also, the loan prediction scenario has been used in other human-AI team studies \citep{Binns2018Loan,Green2019Loan,Gomez2020ViCELoan,Chromik2021Loan,VanBerkel2021Loan,He2023AnalogyExplanations}, reinforcing its validity and suitability for collaboratively analyzing interactions between humans and AI systems.
We decided to convert the nature of this task from low-stakes to high-stakes by rewarding participants with a monetary bonus in case of correct decisions \citep{Salimzadeh2023TaskComplexityDecisionMaking} (see Section \ref{sec:procedure}). Before training the model, we discarded the Loan-ID column given its low informativeness for both the user and the AI in the decision-making process, resulting in \textit{eleven features} (excluding the outcome of the loan request, see Figure \ref{fig:conditions}-A).

\subsubsection{\textbf{Model}}
\label{sec:model_}
We used a Random Forest Classifier (RFC) to solve the loan approval task, following the approach in \citep{Chromik2021Loan}. The RFC was trained with 100 estimators (trees) using an 80:20 stratified split for training and test sets, achieving a test set accuracy of about 83\%, consistent with their results.
We then proceeded to the RFC calibration phase \citep{SilvaFilho2023Calibration} although the methods we tested did not significantly improve the calibration metrics (see Section \ref{sec:model_calib}). We computed the model confidence estimates on the test set, as described in Section \ref{sec:instances_}.
From now on, we will refer to the RFC model as the AI. 

\subsubsection{\textbf{Instances}} 
\label{sec:instances_}
Before selecting the instances for the user study, we computed the AI confidence estimates on the test set using Shannon's entropy method to extract the epistemic uncertainty \citep{Shaker2020RFUncertainties} and convert it into a confidence score ranging from 0 to 100. We computed the quartiles on the test set confidence scores assigning an instance to a low confidence if its value was below 44.3 $(Q_2)$ and a high confidence if its value was above 61.6  $(Q_3)$.
Then, we selected the final instances to include in the user study by picking 16 random instances and balancing them across AI correctness, confidence, predicted class, and true class (see Table \ref{tab:ai_conditions}). Next, we randomly split these instances into two groups of eight, balancing the values of the aforementioned attributes (i.e., our controlled variables). We keep the first group for practice and the latter for the main session. The final low confidence values were between 9\% and 43\%, while
high confidence values were between 68\% and 85\%. Given the test accuracy of the AI is about 83\%, participants ``observed'' accuracy will be only 62.5\%. We deliberately presented more instances where the AI made incorrect predictions to investigate whether and how participants would tend to rely excessively on the AI system.
To account for ordering effects \citep{Nourani2021AnchoringBias}, we prepared 400 random permutations for the practice and main session instances, ensuring each participant sees differently ordered loan requests.

\begin{table}[!h]
    \centering
    \small
    \setlength{\tabcolsep}{4.5pt}
    \caption{Instances settings for practice and main sessions of the loan prediction tasks, for which the order has been uniquely randomized for each participant.}
    
    \begin{tabular}{ccccc}
        \toprule
        \textbf{ID} &
        \textbf{AI correctness} & \textbf{AI confidence} & \textbf{AI prediction} & \textbf{True prediction} \\
        
        \midrule
        1 & correct  & high & reject & reject \\ 
        2 & correct  & low & reject & reject \\  
        3 & wrong  & high & reject & accept \\
        4 & correct  & low & accept & accept \\
        %\midrule
        5 & correct  & high & accept & accept \\ 
        6 & correct  & low & accept & accept \\ 
        7 & wrong  & high & accept & reject \\ 
        8 & wrong  & low & accept & reject \\ 
        \bottomrule
    \end{tabular}
    \label{tab:ai_conditions}
\end{table}

\subsubsection{\textbf{AI assistance and explanations}} 
\begin{figure}[h!]
      \centering
    \includegraphics[width=\textwidth]{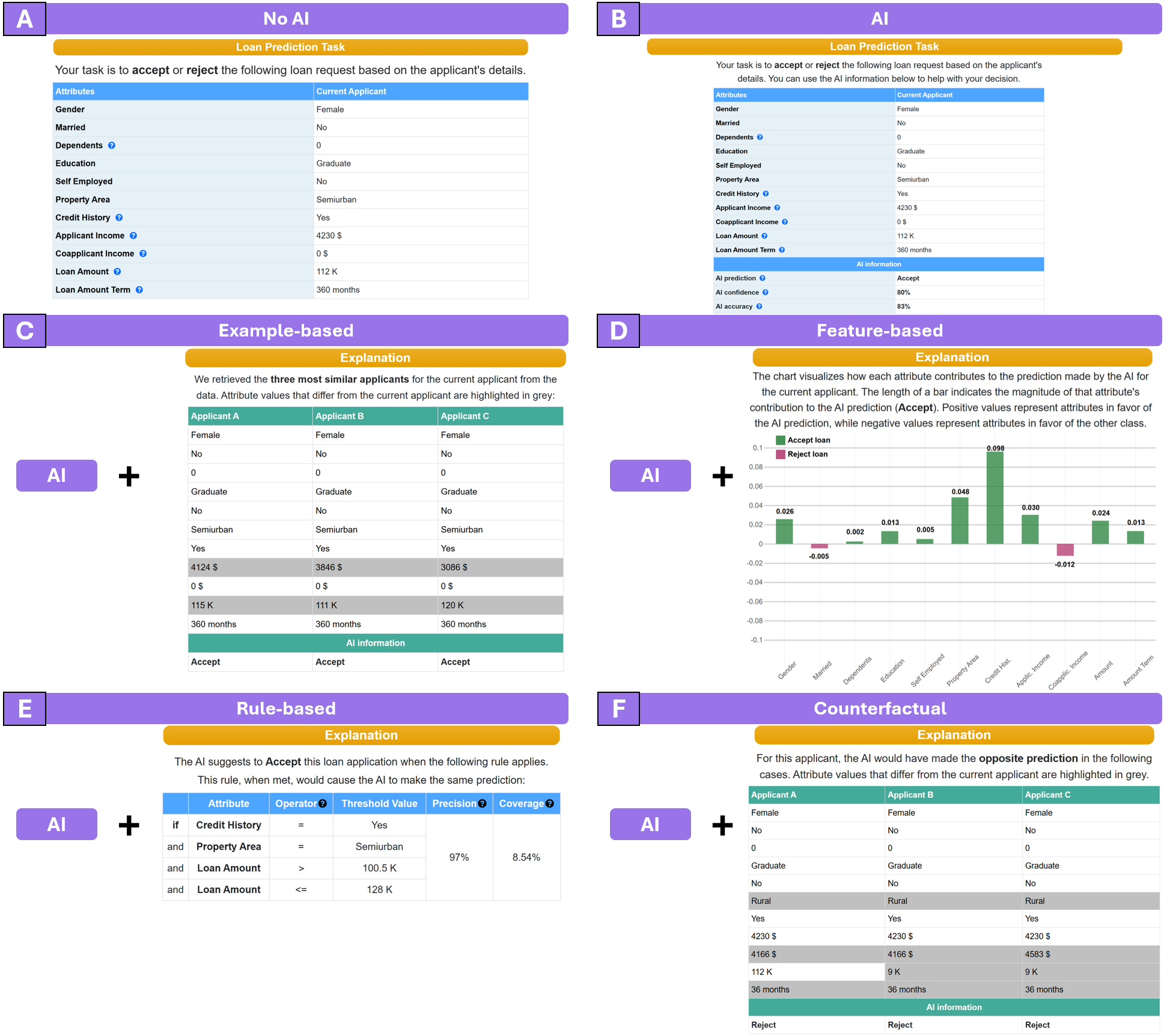}
    \caption{AI assistance conditions for the loan approval tasks. Participants can display additional information about the attributes by hovering over the info buttons. \textbf{(A - No AI)} Participants will see the task's goal and the current applicant's details. \textbf{(B - AI)} Participants will also be assisted by an AI in the decision-making task (i.e., with prediction, confidence, and accuracy). \textbf{(C - Example-based)} Participants will see condition ``B - AI'' and the three nearest neighbors of the current applicant.  
    \textbf{(D - Feature-based)}  Participants will see condition ``B - AI'' and the Shapley feature contribution for each applicant's attribute.
    \textbf{(E - Rule-based)} Participants will see condition ``B - AI'' and the rule generated by Anchor.
    \textbf{(F - Counterfactual)} Participants will see condition ``B - AI'' and three counterfactual instances generated by DiCE. 
    }
    \label{fig:conditions}

\end{figure}

In this work, we assessed the effects of six AI assistance conditions (see Figure \ref{fig:conditions}), using no AI assistance as a baseline. One condition included AI information without explanations, incorporating prediction, confidence in the prediction, and AI accuracy on the test set. The remaining four conditions added explanations to this AI information, as detailed below.

\textit{Example-based.} 
Example-based explanations do not usually provide direct insights into the internal model functioning in predicting a specific output.
Instead, they are usually employed to show representative prototypes of the AI's predicted class or select similar examples \citep{Binns2018Loan,CCAI2019ExampleBased,Dodge2019ExampleBased,Lai2019ExampleBasedFeatureBasedAIPredictions,Bucinca2020ProxyTasks,Hase2020ExampleBased,Wang2021ExampleBasedFeatureBasedAndOthers,Avidan2022ExampleBased} that resemble the examined instance. An exception of this concerns approximating a black-box model to a surrogate transparent model (i.e., Twin Systems \citep{Kenny2019TwinSystems,Kenny2021TwinSystems,Ford2023ImageClassificationPeopleExpertiseExampleBasedExplanations}), where the weights of a black-box model are transferred into a transparent surrogate such as a k-NN. This way, the surrogate model mimics the original black-box model behavior and provides nearest-neighbor instances that align with the original model decisions. In our study, we built example-based explanations taking inspiration from  \citep{Chen2023RelianceExampleBasedFeatureBased}. We selected the three nearest neighbor instances from the training set with the closest standardized Euclidean distance to the current loan request test instance, showing the AI prediction of the neighbor instances. To reduce the cognitive load on users, we highlight the neighbor feature values that differ from the given loan request test instance, so that users can focus on the differences between instances (see Fig. \ref{fig:conditions}-C, Example-based). 

\textit{Feature-based.} Feature contribution enables users to identify the key attributes that significantly influence the AI's output, facilitating informed decision-making and understanding of the AI's behavior (e.g., LIME \citep{Ribeiro2016LIME} and SHAP \citep{Lundberg2017SHAP}).
Given its solid theoretical background, and the faithfulness and robustness in the generated explanations \citep{Bodria2023Benchmarking,Feldkamp2023SHAPRobustness}, we rendered feature-based explanations using the SHapley Additive exPlanations (SHAP) model-agnostic method \citep{Lundberg2017SHAP}, explaining the AI’s prediction by showing the Shapley contribution of each feature in favor (positive sign) or against (negative sign) the AI's prediction and presented with an interactive vertical bar chart (see Fig. \ref{fig:conditions}-D, Feature-based). We used purple to represent contributions of a rejected loan request and green for an accepted loan request. The length of each bar indicates the magnitude of that attribute's contribution relative to the AI prediction on the current loan request.

\textit{Rule-based.} 
Rule-based explanations provide series of \textit{``if-then''} statements highlighting a model's decision-making process that humans can easily understand \citep{Abadi2018XAISurvey,Wang2019Rules,RibeiroAnchors2018,Bodria2023Benchmarking}. 
We generated rule-based explanations via the model-agnostic method called \textit{Anchors} \citep{RibeiroAnchors2018}, which defines a rule (set of predicates) so that an instance is assigned to a specific class only if all its predicates (i.e., features tested with threshold values) satisfy that rule with a high probability. Anchors also return the precision and the coverage of the extracted rule. The precision indicates the quality an anchor predicts the model's output. A high precision value suggests that the anchor is a good predictor of the output variable, while a low precision value highlights that the anchor is a poor predictor. Instead, coverage measures how many examples in the dataset are covered by the anchor. 
A high coverage value indicates that the anchor is a good representative of the dataset, while a low coverage value means the anchor is a poor representative. When generating the rules, we set the precision threshold constraint to 95\% (i.e., finding the anchor that maximizes the coverage given the threshold). We show participants the extracted rule in a tabular form where each row represents a predicate where a feature is tested against a threshold value. Additionally, we added two columns showing the precision and coverage of the generated rule (see Fig. \ref{fig:conditions}-E, Rule-based). 

\textit{Counterfactual.}
Counterfactual explanations provide contrastive \textit{``what-if''} statements that help users understand what changes could be made to achieve a desired output \citep{Wachter2017CounterfactualEW,Abadi2018XAISurvey,Mothilal2020Counterfactuals}.
We built counterfactual explanations using the Diverse
Counterfactual Explanations (DiCE) framework \citep{Mothilal2020DiCE} for its effectiveness in providing diverse and actionable counterfactual explanations \citep{Mothilal2021towardsDice,Moreira2022BenchmarkingCADice}.
Given a test instance, DiCE generates counterfactual explanations that emphasize diversity and deliver a more comprehensive understanding of the model's behavior, providing multiple counterfactuals that are diverse in terms of the changes made to the input features.  
Following the line of example-based explanations, we show users three counterfactual explanations generated from a given loan request test instance. Similarly, we highlight the counterfactual feature values that differ from the given loan request test instance to reduce users' cognitive load and let them focus on the differences between instances (see Fig. \ref{fig:conditions}-F, Counterfactual). 

%% Methods
\section{Study Design}
\label{sec:materials}
Our study followed a mixed-factorial design, where we asked participants to decide whether to accept or reject a series of loan requests (see Table \ref{tab:ai_conditions}). We initially measured participants' NFC and divided them into low and high groups based on the distribution median. Next, we assigned each participant to one of the \textit{AI assistance} conditions as a between-subjects factor (No AI, AI, example-based, feature-based, rule-based, and counterfactual). Also, we studied the effects of the following within-subjects covariates: \textit{AI confidence} (low and high), and \textit{AI correctness} (correct and wrong). 
First, participants completed a practice session of eight loan requests to familiarize themselves with the task and the assigned AI assistance condition. Next, they completed the main session of the study with another eight loan requests. 

This section outlines the variables, planned sample size, statistical analysis, and the procedure for the user study we conducted to test our hypotheses. 

%% Variables
\subsection{Variables}
\label{sec:variables}
For the hypotheses test, we considered the following measurements collected in the \textit{main session} of the user study.
We collected the following independent variables:

\begin{itemize}[]
    \item \textit{AI assistance} (between-subjects, categorical).
    We created six scenarios that varied in terms of assistance provided by the AI and explanations to the participants during their decision-making process.

    \begin{itemize}
        \item \textit{No AI.} 
        We showed participants the loan request attributes and asked whether it should be accepted or rejected.

        \item \textit{AI.} We showed participants the information in the \textit{No AI} condition and the following AI information: i) prediction for the current loan request,  ii) prediction confidence, and iii)  accuracy on the test set.  

        \item \textit{Example-based.} We showed participants the information in the \textit{AI} condition and three nearest neighbor instances of the current loan request.

        \item \textit{Feature-based.} We showed participants the information in the \textit{AI} condition and the SHAP feature contribution for each loan request attribute.

        \item \textit{Rule-based.} We showed participants the information in the \textit{AI} condition and the Anchor rule for the current loan request. 

        \item \textit{Counterfactual.} We showed participants the information in the \textit{AI} condition and three DiCE-generated counterfactual instances based on the current loan request.
        
    \end{itemize}

    \item \textit{Need for Cognition} (between-subjects, categorical). NFC is a stable personality
    trait that reflects how much a person enjoys engaging in cognitively demanding activities \citep{Cacioppo1984NFC}.
    We measured participants' NFC using the six-item Need for Cognition Scale (NCS-6) defined in \citep{LinsDeHolandaCoelho2020NFC6} (see Section \ref{sec:appendix} for details). We split participants into low and high NFC by computing the \textit{median} of the NFC score distribution, the same criteria used in \citep{Bucinca2021NFC}.

\end{itemize}

We measured their effects on four dependent variables:

\begin{itemize}[]
    \item \textit{User accuracy} (categorical). We measured participants' accuracy by assessing whether the decision of a participant to accept or reject a loan aligned with the true loan prediction (i.e., wrong or correct).
    
    \item \textit{Reliance} (categorical). We measured participants' reliance on AI by assessing whether a participant agreed or disagreed with the AI prediction (i.e., agree or disagree).

    \item \textit{Interface components importance} (ranking). We measured the importance of interface elements for participants in determining their final choice, including the loan request, the AI information, and the explanation, measured as a ranking. Participants responded to the statement:
    ``Please rank the following information in terms of how important it was for you in making your final decision: a) loan attributes, b) AI information, c) explanation''.

    \item \textit{Cognitive load} (numerical). We assessed how difficult participants found the tasks using the Single Ease Question (SEQ) \citep{Sauro2009SEQ} 7-point rating scale, ranging from ``1 - Very easy'' to ``7 - Very difficult''. 

\end{itemize}

We also collected the following \textit{covariates} (see Table \ref{tab:ai_conditions}):

\begin{itemize}[]

     \item \textit{AI confidence} (within-subjects, categorical). Participants saw loan requests with either low or high AI confidence.

     \item \textit{AI correctness} (within-subjects, categorical). Participants saw loan requests with correct or wrong AI predictions. 
\end{itemize}

Finally, we collected other descriptive and exploratory measurements to provide context for our study and enable further exploratory analyses to motivate our hypotheses: 

\begin{itemize}[]
     \item \textit{Demographics} (categorical). We gathered participants' information on their sex and age from the Prolific platform. 
     
     \item \textit{Familiarity with the task} (categorical). We asked participants about their familiarity with loan request approval with the following statements using a 5-point Likert scale ranging from ``1 - No experience'' to ``5 - Highly experienced'':
     \begin{itemize}
         \item ``Do you have any experience with loan request approval?”

         \item ``Do you have any experience with AI-assisted loan request approval?”
\end{itemize}

     \item \textit{AI information importance} (ranking). 
     We asked participants to rank the importance of the AI prediction, confidence, and accuracy in the conditions that include the AI information by asking:  ``Please rank the following AI information in terms of how important it was for you in making your final decision:  a) AI prediction b) AI confidence, c) AI accuracy''.

    \item \textit{XAI interface understanding} (numerical). At the end of the survey, we asked participants to state their easiness of understanding the loan application attributes, AI information, and explanations using a 5-point Likert scale ranging from “1 - Strongly disagree” to “5 - Strongly agree” in three items (i) ``The loan application attributes were easy to understand'', (ii) `The AI information provided was easy to understand'', and (iii) ``The AI explanation provided was easy to understand''. 

     \item \textit{Textual feedback} (open text). At the end of the survey, we collected participants' feedback about the explanations (when presented) by asking: ``What were the pros and cons of the AI explanations you encountered?'' 
  
\end{itemize}

\subsection{Planned Sample Size and Statistical Analysis}
\label{sec:participants}
Before recruiting participants, we estimated the required sample size for our study using \textit{G*Power} software \citep{faul2009statistical}, resulting in 286 participants. This recommended sample size is motivated by the maximum number of participants needed among hypotheses, which we describe in detail as follows.
Since we are assessing five hypotheses with mixed models (continuous/categorical dependent variables) and two based on ranking information (using the Friedman test), we decided to apply two different thresholds, using $\alpha=\frac{0.05}{5}=.01$ for mixed models and $\alpha=\frac{0.05}{2}=.025$ ranking tests.
Thus, we considered as significant the \textit{p}-values below these reduced thresholds in the analysis. Additionally, we assigned a randomly generated seed to each user as a (i) random intercept to account for the variability of the dependent variables across different clusters in the mixed-effects logistic regression and as a (ii) within-cluster correlation effects on the dependent variable in the Generalized Estimation Equation (GEE) models. All the models converged successfully. 

To answer H1a and H1c with categorical dependent variables, we used two mixed-effects logistic regression models with \textit{Reliance}  and \textit{User accuracy}  as the dependent variables, assessing the main effects of \textit{AI assistance}  as the independent variable, and \textit{AI confidence} and \textit{AI correctness} as covariates. We computed the required sample size using \textit{G*Power} for a mixed-effects logistic regression model (a priori ${\chi}^2$ test) with medium effect size (Cohen's $d$ = 0.25), a desired power of $0.8$, Df = 5, and two covariates (AI confidence and AI correctness), resulting in 286 participants\footnote{While H1a and H1b require around 191 participants (Df=1) for low and high AI confidence levels, H1c increases the number of participants given that we tested all six AI assistance conditions (Df=5).}. Instead, to answer H1b which involves a numeric dependent variable, we used a Generalized Estimation Equation (GEE) model with \textit{Cognitive load} as the dependent variable to assess the main effects of the \textit{AI confidence} covariate while also studying potential impacts of the \textit{AI assistance} as an independent variable and \textit{AI correctness} as a covariate. We computed the required sample size using the \textit{G*Power} for a mixed-design ANCOVA, medium effect size (Cohen's $f$ = 0.25), a desired power of $0.8$, Df=1, and two covariates (AI confidence and AI correctness), resulting in 191 participants.

To answer H2, we conducted a Friedman test \citep{friedman1,friedman2} with \textit{Interface component importance} ranked measurements as the dependent variable to assess the main and interaction effects of \textit{Need for Cognition} (low and high) as the independent variable. We computed the required sample size using \textit{G*Power} for a within-subjects Friedman Test with medium effect size (Cohen's $f$ = 0.16), a desired power of $0.8$, one group, and three measurements (i.e., loan application attributes, AI information, and explanation), resulting in 100 participants. To establish the ranking order among XAI interface elements, we conducted a Nemenyi posthoc analysis when we discovered significant
factors in the Friedman test.

To answer hypothesis H3a with a categorical dependent variable, we used a mixed-effects logistic regression model with \textit{User accuracy} as the dependent variable to study the main effects of \textit{Need for Cognition} independent variable. We also investigated the impact of \textit{AI assistance} as an independent variable, and \textit{AI confidence} and \textit{AI correctness} as covariates. We computed the required sample size using the \textit{G*Power} for a mixed-effects logistic regression model (a priori ${\chi}^2$ test) with medium effect size (Cohen's $d$ = 0.25), a desired power of $0.8$, Df=1, and two covariates (AI confidence and AI correctness) resulting in 187 participants.
Instead, to answer H3b which involves a numeric dependent variable, we used a Generalized Estimation Equation (GEE) model with \textit{Cognitive load} as the dependent variable to assess the main effects of \textit{Need for Cognition}. Further, we also investigated the impact of \textit{AI assistance} as an independent variable, and \textit{AI confidence} and \textit{AI correctness} as covariates. We computed the required sample size using the \textit{G*Power} for a mixed-design ANCOVA, medium effect size (Cohen's $f$ = 0.25), a desired power of $0.8$, Df=1, and two covariates (AI confidence and AI correctness), resulting in 191 participants.

\subsection{Procedure}
\label{sec:procedure}
To verify our hypotheses, we conducted an online user study using the Prolific platform\footnote{\href{https://www.prolific.com/ }{https://www.prolific.com/} \label{prolific}}, where we recruited participants aged 18 or older with high English proficiency and approval rates between 95 and 100. Participants were then redirected to the LimeSurvey tool\footnote{\href{https://www.limesurvey.org/ }{https://www.limesurvey.org/} \label{limesurvey}} where they completed the study in three steps. 
Participants received \pounds 2.7 as a reward for the study, with an average completion time of 18 minutes (i.e., \pounds 9/hour, which is considered a fair payment for Prolific). Prolific automatically timed out participants after 60 minutes.
We rewarded participants with an extra \pounds 0.12 for each correctly classified loan request of the main session. We only included participants in the analysis if they passed all five attention checks. The study has been approved by the Ethics Committee of the University of Cagliari\footnote{Received on 25 July 2024, Prot. 0205640.}.

\begin{figure}
      \centering
    \includegraphics[width=\textwidth]{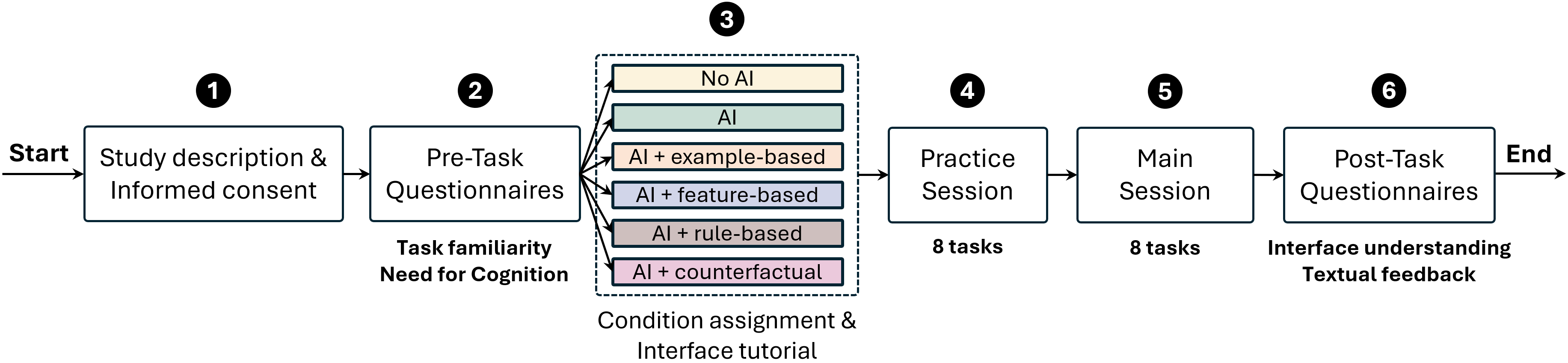}
    \caption{Illustration of the procedure participants engaged in during our study. 
}

    \label{fig:procedure}
\end{figure}

Participants went through the following steps, illustrated in Figure \ref{fig:procedure}. 
First, they read a document containing a brief study description, filled out an informed consent form, and completed an attention check\footnote{We use Instructional Manipulation Checks (IMCs), where the answer to each attention check is explicitly reported in the question text and follows the good practices of Prolific \href{https://researcher-help.prolific.com/en/article/fb63bb}{(see link)}. \label{attention_checks}}. 
Next, they stated their familiarity with the task and completed another attention check. Then, we asked participants to fill out the six-item Need for Cognition Scale \citep{LinsDeHolandaCoelho2020NFC6} and to complete another attention check.
We introduced participants to the task and assigned them to one of the six AI assistance conditions (i.e., \textit{No AI}, \textit{AI}, \textit{Example-based}, \textit{Feature-based}, \textit{Rule-based}, and \textit{Counterfactual}) while balancing the participation among conditions. 
Before starting the practice session, participants completed another attention check. Then, participants completed eight loan request tasks as a practice session, where they needed to decide whether to accept or reject the applications.
After each decision, participants \textit{received feedback} on their answers, where we reveal the corresponding true class.
When participants finished the practice session, we showed them a page as a reminder for the main task session resulting in a compensation bonus in case of correctly classifying a loan. Before starting the main session, participants completed the last attention check. 

Participants completed eight loan request tasks, with the same AI assistance condition assigned in Step 2 but \textit{without receiving feedback} on the true class. For each task, we measured participants' cognitive load. We also asked them to rank the importance of the interface components (see Section \ref{sec:variables}) except in the \textit{``No AI''} and \textit{``AI''} conditions. 
Finally, we asked participants to state their easiness of
understanding of the XAI interface elements (i.e., loan application attributes, AI information, and explanation), and to provide textual feedback about the pros and cons of the explanations they encountered (see Section \ref{sec:materials}). 

\section{Results}
\label{sec:results}
\subsection{Descriptive Statistics}

The final sample of 288
participants comprised 144 males and 144 females, aged between 18 and 74 ($M$ = 32.42, $SD$ = 10.95).
Participants reported low familiarity with the loan application task ($M$ = 1.83, $SD$ = 0.99, 5-point Likert scale, 1: no experience, 5: highly experienced) and AI-assisted loan request approval ($M$ = 1.32, $SD$ = 0.71, 5-point Likert scale, 1: no experience, 5: highly experienced).
Overall, participants reported a good easiness in understanding the loan application attributes ($M$ = 3.72, $SD$ = 0.93, 5-point Likert scale, 1: strongly disagree, 5: strongly agree), AI information ($M$ = 3.74, $SD$ = 0.95, 5-point Likert scale, 1: strongly disagree, 5: strongly agree), and explanations ($M$ = 3.67, $SD$ = 1.00, 5-point Likert scale, 1: strongly disagree, 5: strongly agree).

\subsection{Hypothesis Tests}
\subsubsection{H1: Effects of AI and explanations on users' reliance on AI, cognitive load, and accuracy} 
\begin{figure}
      \centering
    \includegraphics[width=\textwidth]{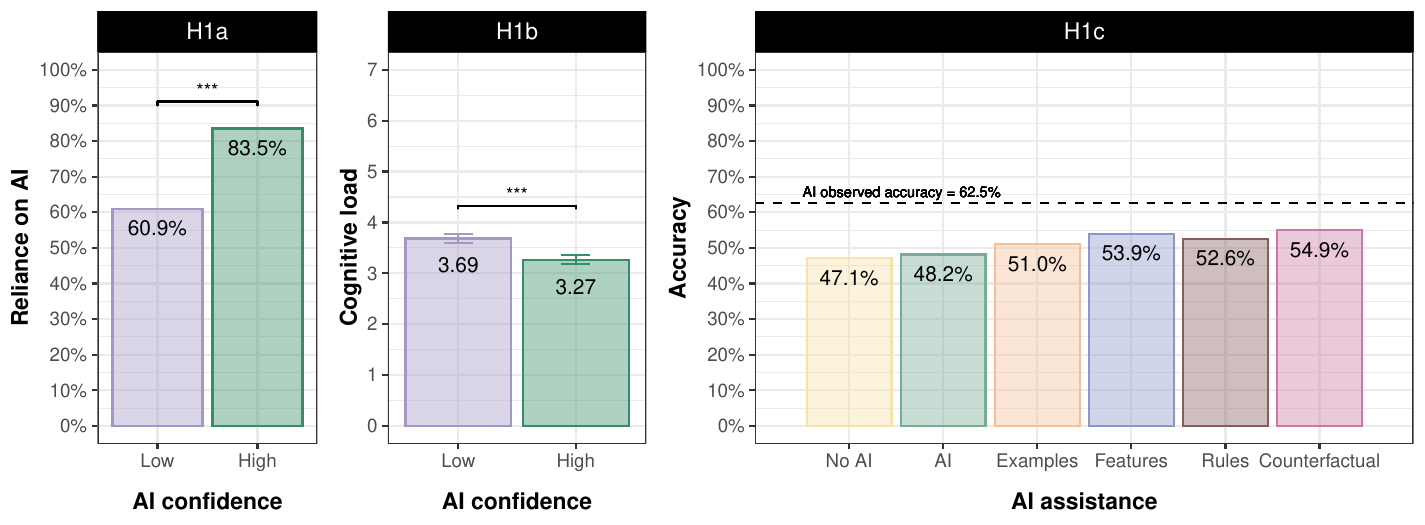}
    \caption{Effects of low and high AI confidence considering reliance on AI \textbf{(H1a)}, cognitive load \textbf{(H1b)} (ticks above bars indicate lower and higher confidence intervals based on standard errors) and users' accuracy \textbf{(H1c)} divided by AI assistance conditions. The asterisks highlight p-value significance strength (***$p$ $<$ .001).
}

    \label{fig:hyp1}
\end{figure}

The resulting charts for H1 are depicted in Figure \ref{fig:hyp1}.
For \textbf{H1a}, we used a mixed-effects logistic regression model to examine the differences in users' reliance on AI considering low and high AI confidence. The results of the analysis showed a significant effect (\textit{Log-Odds} = 1.22, \textit{Std. error} =  0.12,  \textit{z-value} = 10.40, $p$ $<$ .01) of high AI confidence in increasing users’ reliance on AI than low AI confidence. Hence we \textit{reject the null hypothesis} for \textbf{H1a}, as users rely more on the AI when exposed to high AI confidence than low confidence. In \textbf{H1b}, we studied the differences in users' cognitive load between low and high AI confidence using a Generalized Estimation Equation (GEE) model. The results of the analysis showed a significant effect (\textit{Log-Odds} = -0.41, \textit{Std. error} =  0.06,  \textit{Wald} = 54.57, $p$ $<$ .01) of high AI confidence in decreasing users' cognitive load compared to low AI confidence. Hence we \textit{reject the null hypothesis} for \textbf{H1b}, concluding that users report lower cognitive load when exposed to high AI confidence compared to low confidence. For \textbf{H1c}, we investigated the users' accuracy differences among AI assistance conditions using a mixed-effects logistic regression model. The results of the analysis showed no significant effects (\textit{Log-Odds} = 0.34, \textit{Std. error} =  0.16,  \textit{z-value} = 2.11, $p$ = .0349) of feature-based explanations over the other interface conditions on users' accuracy, hence we \textit{fail to reject the null hypothesis} for \textbf{H1c} \footnote{Although the result did not meet the $\alpha = .01$ threshold, counterfactual explanations were the only other explanation type, besides feature-based explanations, to show an effect on improving users' accuracy (\textit{Log-Odds} = 0.39, \textit{Std. Error} = 0.16, \textit{z} = 2.43, $p$ = .0149).}.

\subsubsection{H2: Effects of low and high NFC participants on XAI interface information importance.}
\begin{figure}
      \centering
    \includegraphics[width=\textwidth]{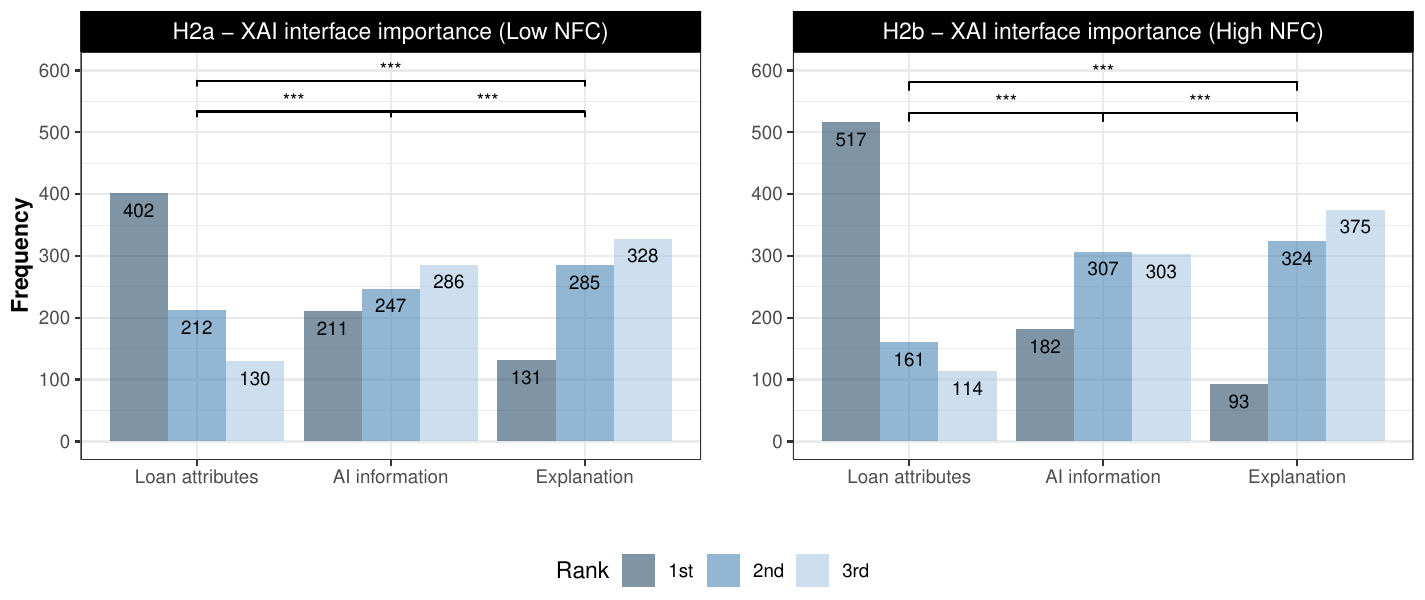}
    \caption{XAI interface components rank frequencies for low \textbf{(H2a)} and high \textbf{(H2b)} NFC individuals. The asterisks highlight p-value significant strength (***$p$ $<$ .001). }

    \label{fig:hyp2}
\end{figure}

\hfill\\
To test H2 (see Figure \ref{fig:hyp2}), we included only participants exposed to explanations, resulting in 192 users. For \textbf{H2a}, we hypothesized that low NFC participants would give priority to the AI information (rank 2) immediately after the loan attributes (rank 1), keeping the explanation (rank 3) as a last resource.
The Friedman test for \textbf{H2a} shows a significant difference (${\chi}^2$ = 159, \textit{df} = 2, \textit{p} $<$ .025) between the three XAI interface elements when investigating low NFC participants. The pairwise ranking comparisons using the Nemenyi (\textit{p} $<$ .025) show that users prioritize the loan attributes (rank 1), followed by the explanation (rank 2) and the AI information (rank 3) when making their final decision. In this light, we \textit{fail to reject the null hypothesis} for \textbf{H2a}.
For \textbf{H2b}, the Friedman test shows a significant difference (${\chi}^2$ = 324, \textit{df} = 2, \textit{p} $<$ .025) between the three XAI interface elements when investigating high NFC participants. The Nemenyi pairwise ranking comparisons (\textit{p} $<$ .025) align with our hypothesis, showing that users prioritize the loan attributes (rank 1), followed by the explanation (rank 2) and the AI information (rank 3) when making their final decision. Hence, we \textit{reject the null hypothesis} for \textbf{H2b}.

\subsubsection{H3: Effects of low and high NFC participants on accuracy and cognitive load.}
\begin{figure}
      \centering
    \includegraphics[width=\textwidth]{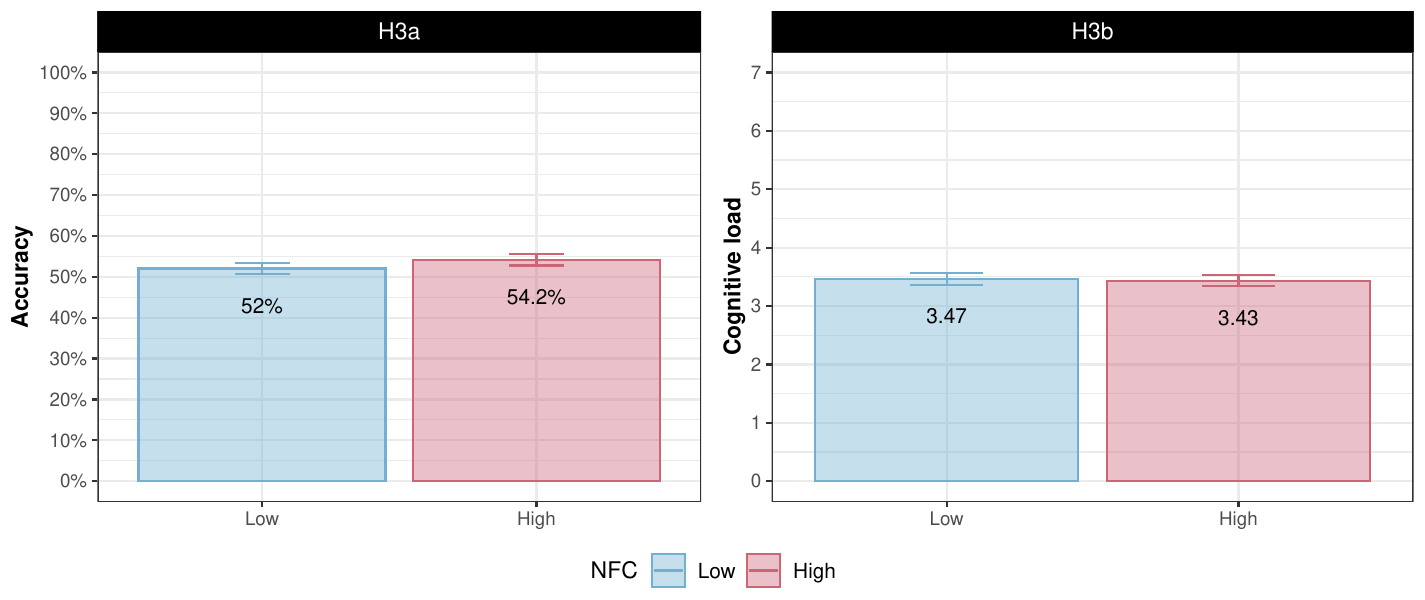}
    \caption{Users' accuracy \textbf{(H3a)} and cognitive load \textbf{(H3b)} disaggregated by low and high NFC (ticks above bars indicate the Standard Error). }

    \label{fig:hyp3}
\end{figure}

\hfill\\
For \textbf{H3a} (see Fig. \ref{fig:hyp3}), we investigated whether users with a high NFC may have an increase in accuracy than users with a low NFC when exposed to explanations. 
The results of the mixed-effects logistic regression analysis showed no significant effects (\textit{Log-Odds} = 0.03, \textit{Std. error} =  0.10,  \textit{z-value} = 0.28, $p$ = .78) among low and high NFC participants. Hence, we \textit{fail to reject the null hypothesis} for \textbf{H3a}. 
In \textbf{H3b}, we studied the differences in users' cognitive load between low and high NFC participants when exposed to explanations using a Generalized Estimation Equation (GEE) model. The results of the analysis showed no significant effects (\textit{Log-Odds} = -0.08, \textit{Std. error} =  0.12,  \textit{Wald} = 0.51, $p$ = .47) for high NFC participants compared to low NFC participants. Hence we \textit{fail to reject the null hypothesis} for \textbf{H3c}. 

%%%%% table for main hypotheses
\begin{table}[!h]
    \centering
    \small
    \setlength{\tabcolsep}{3pt}
    \caption{Summary results of our hypotheses.}
    \begin{tabular}{p{10.5cm} P{2cm}}
        \toprule
        \textbf{Hypotheses} & \textbf{Supported}  \\
        
        \midrule
        \textbf{H1a}: Users exposed to a high AI confidence will rely more on the AI prediction than users exposed to a low AI confidence & \cmark \\ 
        
        \textbf{H1b}: Users exposed to a high AI confidence will report a lower cognitive load than users exposed to a low AI confidence
         & \cmark \\  

         \textbf{H1c}:  Users exposed to feature-based explanations will achieve higher accuracy than other AI assistance conditions
         & \xmark \\ 

         \midrule

         \textbf{H2a}: Users with a low NFC will mainly prioritize the applicant's details to make their final decision (rank 1), then the AI information (rank 2), and lastly the explanation (rank 3) 
            & \xmark \\ 
        
        \textbf{H2b}: Users with a high NFC will mainly prioritize the applicant's details to make their final decision (rank 1), then the explanation (rank 2), and lastly the AI information (rank 3) 
        & \cmark \\

        \midrule

         \textbf{H3a}: When explanations are shown, users with a high NFC will achieve a higher accuracy than users with a low NFC  & \xmark \\ 
        
        \textbf{H3b}: When explanations are shown, users with a high NFC will report a lower cognitive load than users with a low NFC        & \xmark \\

        \bottomrule
    \end{tabular}
    \label{tab:main_hyp}
\end{table}

\subsection{Post Hoc and Exploratory Analyses}
\label{sec:posthoc}
The hypotheses results (see Table \ref{tab:main_hyp}) revealed that high AI confidence increases reliance on AI and reduces cognitive load. Additionally, there were no significant differences in user accuracy among the different AI assistance conditions. Considering the interface component preferences, low and high NFC participants ranked loan attributes first, explanation second, and AI information third. 
Finally, no accuracy or cognitive load differences between low and high NFC individuals were found.

To further clarify the role of AI and explanations in shaping user behavior, we conducted additional analyses considering the interaction effects between covariates (AI confidence and correctness) and explanations, further clarifying the role of AI information in users' prioritization of XAI interface elements' ranking. 
We first examined how AI confidence influences users' interpretation of explanations by considering metrics such as accuracy, reliance on AI, and cognitive load. We then reassessed these metrics by considering AI correctness to investigate potential overreliance behavior in AI when users interact with explanations. 
Additionally, given the significant impact of high AI confidence on increasing users' reliance on AI, we evaluated how it impacted users' prioritization of the XAI interface elements (i.e., loan attributes, AI information, and explanation) and whether it affected users' ranking of AI information (i.e., prediction, confidence, and accuracy).
Lastly, we focused on how low and high NFC users ranked the AI information (i.e., prediction, confidence, and accuracy), where we considered only the AI assistance condition incorporating explanations.  

\begin{figure}
      \centering
    \includegraphics[width=\textwidth]{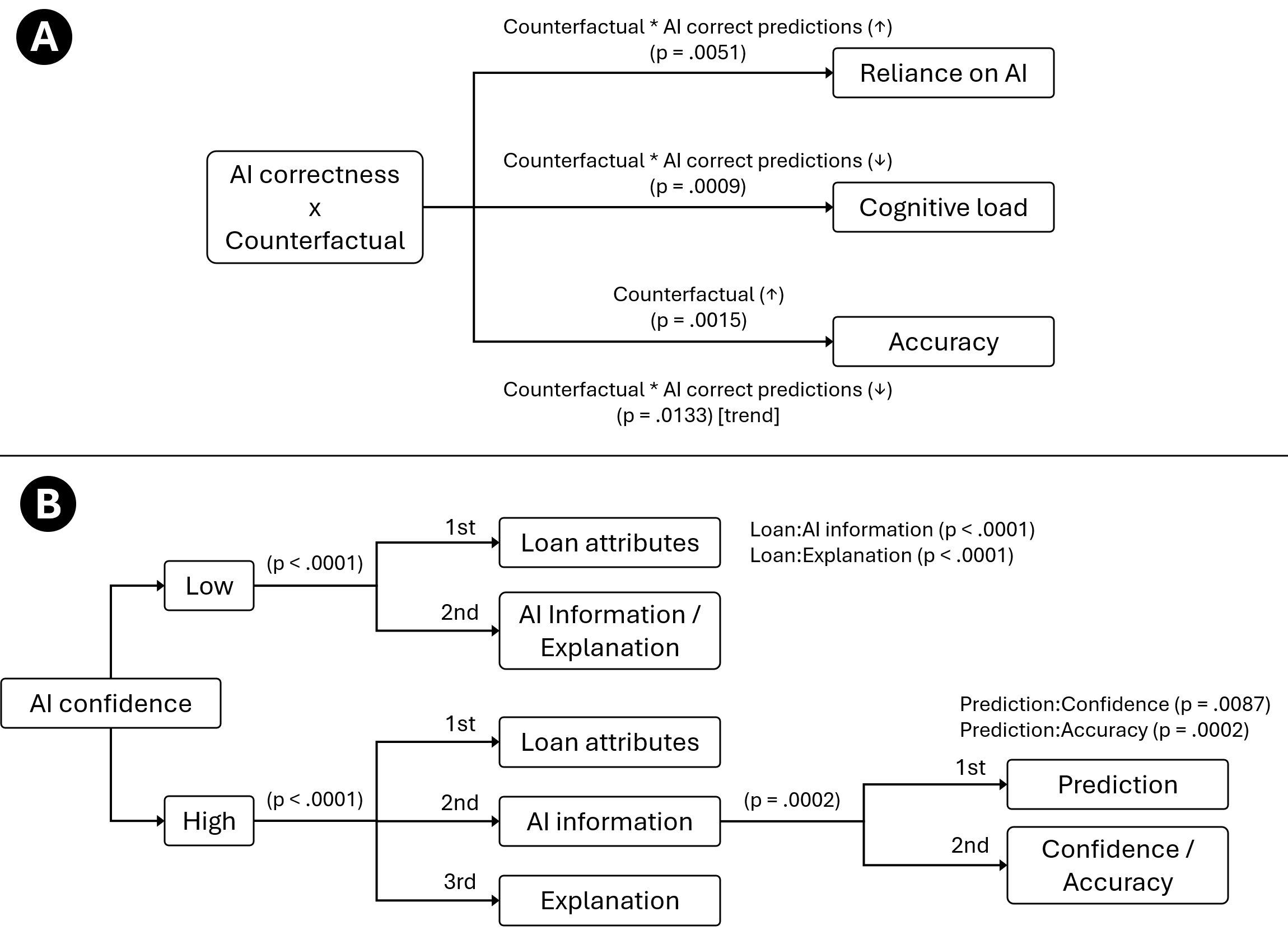}
    \caption{Post hoc analyses results for (A) AI correctness interaction with AI assistance, and (B) ranking for low and high AI confidence with AI information importance of interface elements. The connections between rows present p-values and the direction of the effect (e.g., a downward arrow for a decrease in the connected dependent variable; for rankings, we display the exact position of each interface element based on pairwise comparisons). }

\label{fig:posthoc}
\end{figure}

 The results from the first analysis show no significant interactions between AI confidence and explanations of users' reliance on AI, cognitive load, and accuracy\footnote{Although it falls outside the scope of our hypotheses, it is important to notice that high AI confidence significantly increases users' accuracy ($p$ $<$ .01).}. 
Instead, we found multiple significant results when considering the AI correctness and explanation interactions (see Fig. \ref{fig:posthoc}-A). For reliance on AI, counterfactual explanation interaction with AI correct predictions leads to an increase in reliance (\textit{Log-Odds} = 0.98, \textit{Std. error} =  0.35,  \textit{z-value} = 2.79, $p$ = .0051).
The cognitive load results for counterfactual explanations and interaction with AI correctness (\textit{Log-Odds} = -0.48, \textit{Std. error} =  0.14,  \textit{Wald} = 10.91, $p$ = .0009) show a decrease in users' cognitive load. 
These findings suggest that presenting counterfactual explanations reduces the cognitive load when AI predictions are correct. Additionally, such explanations encourage users to follow correct predictions, potentially mitigating overreliance on AI.

Interestingly, users' accuracy findings highlight a trend for AI correct predictions interacting with counterfactual explanations (\textit{Log-Odds} = -0.84, \textit{Std. error} =  0.34,  \textit{z-value} = -2.47, $p$ $<$ .0133) in decreasing accuracy. Additionally, counterfactual explanations  (\textit{Log-Odds} = 0.87, \textit{Std. error} =  0.27,  \textit{z-value} = 3.17, $p$ = .0015) lead to an increase in accuracy. These results might indicate a nuanced trade-off: counterfactual explanations improve decision-making overall but can sometimes confuse users when AI predictions are already correct.

The results of splitting XAI interface information by AI confidence (see Fig. \ref{fig:posthoc}-B) show a significant difference between the three interface components for low confidence
 (${\chi}^2$ = 301, \textit{df} = 2, \textit{p} $<$ .025). The Nemenyi pairwise comparisons show a significant difference (\textit{p} $<$ .025) between loan attributes (rank 1) with AI information and explanation. Instead, there are no differences among AI information and explanation. We also have a significant difference among the three interface components for high AI confidence (${\chi}^2$ = 196, \textit{df} = 2, \textit{p} $<$ .025). The Nemenyi pairwise comparison results (p $<$ .025) show that participants prioritize the loan attributes (rank 1), followed by the AI information (rank 2), and then the explanation (rank 3). Finally, we found no ranking differences among AI prediction, confidence, and accuracy when considering low AI confidence. Instead, the results for high AI confidence highlight a difference among the AI information elements (${\chi}^2$ = 17.3, \textit{df} = 2, $p$ $<$ .025). The Nemenyi pairwise comparisons ($p$ $<$ .025)
reveal a significant difference between AI prediction and both AI confidence and accuracy, while no significant difference is observed between AI confidence and accuracy.

In the second analysis, we repeated the Friedman test focusing on the AI prediction, confidence, and accuracy ranking considering low and high-NFC participants. The results for low NFC participants show a significant difference between AI information provided (${\chi}^2$ = 13.2, \textit{df} = 2, \textit{p} $<$ .025). The Nemenyi pairwise comparisons (\textit{p} $<$ .025) reveal a significant difference between AI prediction over AI accuracy. However, no differences emerge considering AI confidence when compared to AI prediction and accuracy, delining the interchangeability of AI confidence over AI prediction and accuracy. Instead, the Friedman test for high NFC participants highlights no significant differences among AI prediction, confidence, and accuracy. This may hint that low NFC users seem to focus more on the AI prediction, which is reinforced by AI confidence, while high NFC people seem to look at the AI information as a whole.

\subsubsection{Participants' Interface Understandability and Qualitative Feedback}
This section summarizes users' understanding of the interface components and textual feedback on explanation types we collected from the user study, highlighting subjective perspectives and perceived pros and cons from users about explanations. 

\begin{figure}
      \centering
    \includegraphics[width=\textwidth]{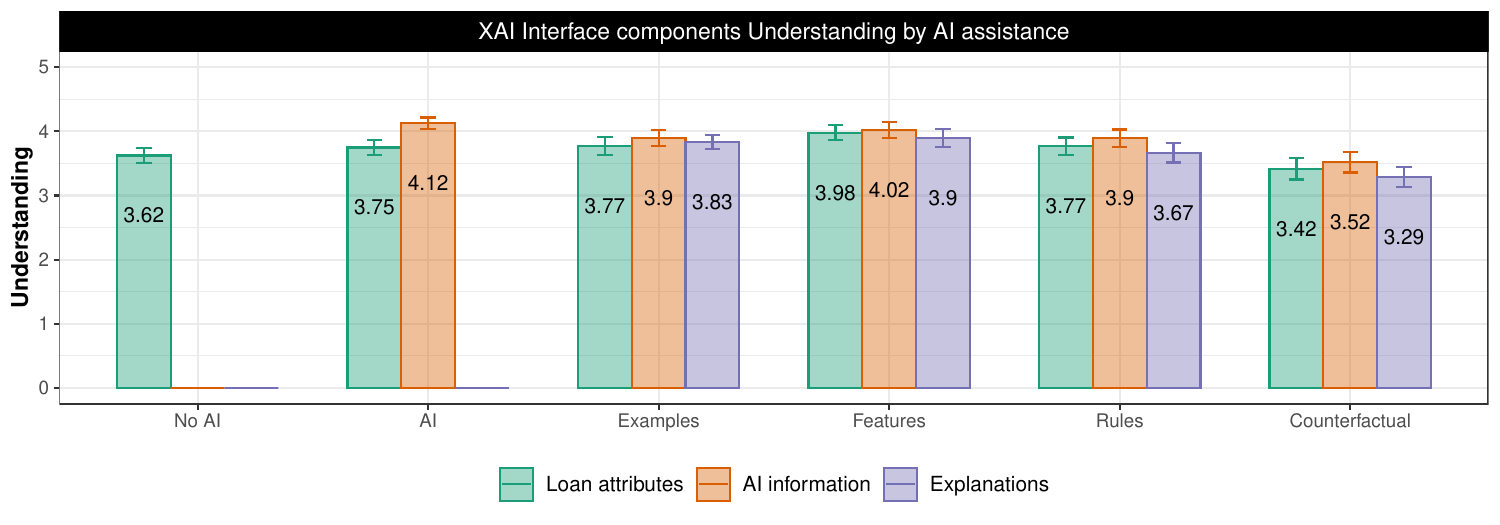}
    \caption{Users' understanding of loan attributes, AI information, and explanations by AI assistance conditions. }
    \label{fig:understanding}
\end{figure}

The chart depicting users' overall understanding of loan attributes, AI information, and explanations is shown in Figure \ref{fig:understanding}.
We notice that, in general, counterfactual explanations decrease overall understanding of interface components. We then conducted a statistical analysis to understand if these differences are merely visual trends or if there is indeed a significant difference.
Given the non-normal nature of the interface components distributions, we opted for a non-parametric Kruskal-Wallis test, using the above variables as dependent variables and the design as the independent variable. Although there were no differences for loan understanding among conditions, we found significant differences for AI (${\chi}^2$ = 9.76, \textit{df} = 4, \textit{p} = .045) and explanation (${\chi}^2$ = 9.92, \textit{df} = 3, \textit{p} = .019) understanding.
We performed a pairwise comparison using a Dunn test with Bonferroni for p-values adjustment. We found a difference between AI and counterfactual conditions (\textit{z} = -2.88, \textit{p} = .0389) for the AI information understanding and another difference among feature-based and counterfactual conditions (\textit{z} = -3.018, \textit{p} = .0152) in the explanation understanding.

Considering users' feedback on explanations, 11 participants reported that example-based ones were easy, understandable, and a fast way to compare applications. As such, P16 said: \textit{\enquote{[explanation] was helpful once understood all the attribute details}}. On the contrary, 11 participants said that explanations lacked details and that it was hard to trust them fully. P73 stated: \textit{\enquote{[explanation] made it easy for making a decision but not sure about their reliability}}.

Feature-based explanations were perceived by 8 participants as helpful and providing clarity for the decision-making. P75 stated: \textit{\enquote{explain well the rationale behind accepting or rejecting the loan}}. However, 10 participants reported needing more insights into why specific weights were assigned to attributes. As such, P79 said: \textit{\enquote{The explanation needed more insights about how the weights were generated}}.

12 participants perceived rule-based explanations as useful and easy to understand, providing good guidance in decision-making. For example, P22 said: \textit{\enquote{The explanation helped me decide whether my evaluation of the loan application is more or less correct or not}}. Despite this, 12 participants stated these explanations lacked understandability, highlighting the absence of \enquote{reasoning} for the rules. As such, P84 reported: \textit{\enquote{Some rules had more information than others which made the choices slightly harder}}.

6 participants perceived counterfactual explanations as helpful and easy to read. For example, P85 reported: \textit{\enquote{The explanation includes many changes in the attribute but helps to understand (going through scenarios) which attributes are more important and influential than others.}}. On the contrary, 6 participants stated they were unclear and trustworthy. For example, P5 said: \textit{\enquote{Explanation is very helpful but hard to trust due to not knowing the mechanisms behind the AI}}.

\section{Discussion}
\label{sec:discussion}
The paper explored how AI assistance and various explanation types influence users' accuracy, reliance on AI, and cognitive load. Additionally, we examined the role of XAI interface elements for individuals with low and high NFC, analyzing differences in accuracy and cognitive load across these groups. Based on our results, we present a comprehensive discussion of our key findings, offering insights into design implications and examining user behaviors in the context of a loan application scenario.

\subsection{The Role of AI in Shaping User Decision-Making}
Our findings reveal that high AI confidence increases users' reliance on AI prediction. This is supported by post hoc analysis, where users prioritize AI information (rank 2) directly after loan attributes (rank 1). Conversely, when exposed to low AI confidence, users prioritize loan attributes over AI information and explanations.
Interestingly, prior research \citep{Cau2023LogicalReasoningStock} in high-uncertainty domains like stock trading found that users prioritize data or AI information interchangeably (rank 1) with high AI confidence, but rank AI (2nd) immediately after data (1st) when AI confidence is low. This suggests that as uncertainty in decision-making increases, individuals are more likely to seek additional guidance from AI. In this context, the confidence level of the AI is essential to the decision-making process.
Our results also indicate that high AI confidence reduces cognitive load, with only a few studies supporting this direction \citep{Souchet24CognitiveLoadHypothesis,SteyversKumar2024CognitiveEffort}. Altogether, our findings reinforce prior work where users tend to rely more on high AI confidence across various domains and tasks \citep{Zhang2020ConfidenceExplanationsAccuracyTrust,Rechkemmer2022AIPerformanceAndConfidenceEffectsOnUsers,Cau2023ExplImageText,Cau2023LogicalReasoningStock,Shuai24SelfConfidenceCalibration,Kahr2023HumanLikeVsAbstractExplanations}. 

While we balanced participants' exposure to low and high AI confidence, they encountered more instances with low confidence and correct predictions than with other combinations of confidence and correctness.
This distribution was intentionally designed to reflect a potential real-world scenario and to study participants' reliance behavior on AI, where the stated AI accuracy (83\%) might not align with the observed accuracy (63\%) on unseen instances.
As summarized in Table \ref{tab:tasks_overview}, users' performance in the loan prediction tasks highlights a clear split between low and high AI confidence instances, particularly considering under-reliance on correct suggestions with low confidence and (over)reliance on wrong suggestions with high confidence. These results highlight the participants' uncertainty in their decision-making and their lack of self-confidence.
Since we can estimate AI confidence but cannot directly control the correctness of predictions for unseen instances, it is essential to explore alternative strategies to optimize the use of AI confidence estimates. 
Consequently, while presenting AI confidence to users is essential for enhancing transparency~\citep{Bertrand2022BiasesInXAIMitigation,Fok2024Aleatoric},  its significant impact on reinforcing AI predictions underscores the need for targeted interface design interventions.

Presenting AI confidence to users is essential for enhancing transparency~\citep{Bertrand2022BiasesInXAIMitigation,Fok2024Aleatoric}; however, its significant influence on reinforcing AI predictions highlights the necessity for targeted interface design interventions. These interventions aim to mitigate users' tendency to over-rely on AI.
AI confidence calibration approaches~\citep{SilvaFilho2023Calibration,Shuai24SelfConfidenceCalibration} provide estimates that accurately reflect the likelihood of correctness in AI predictions. Therefore, it is important to cultivate user awareness regarding their own decision confidence and to determine strategically when to present AI suggestions based on both user and AI confidence levels. One possible solution is to calibrate users' confidence without initial AI assistance, allowing them to receive feedback on the trade-offs between their confidence and accuracy. Once users have developed their confidence, AI assistance can be introduced using design patterns that accommodate both one-stage and two-stage decision-making processes.
For instance, research~\citep{Shuai2023CorrectnessLikelihoodAIUsers,Shuai24SelfConfidenceCalibration} suggests dynamically adjusting the timing of AI assistance by comparing the confidence levels of the user and the AI. AI advice may be omitted or provided on-demand~\citep{Bucinca2020ProxyTasks} when user confidence is high, thereby preserving user autonomy. Conversely, when AI confidence is higher, suggestions can be presented before users make their decisions. This approach balances optimizing AI support with the maintenance of users' agency.

\subsection{The Impact of Explanation Types on User Behavior}
In line with previous studies on the effects of explanations on users \citep{Zhang2020ConfidenceExplanationsAccuracyTrust,Chen2023RelianceExampleBasedFeatureBased,Celar2023CounterfactualCausal}, our results showed that the feature-based explanation might not improve accuracy compared to the other AI assistance conditions. The counterfactual was the only type of explanation closest to our threshold in increasing the accuracy of users, although we did not find differences among the other AI conditions.
The post hoc analysis highlights multiple benefits for counterfactual explanations: increasing users' reliance on AI while diminishing cognitive load when correct AI predictions are shown, and potentially increasing accuracy. Nevertheless, a trend suggests they might occasionally lower accuracy in specific contexts (correct AI predictions) and be perceived as less understandable, as highlighted by our qualitative analysis. 
Interestingly, despite having nearly identical visualizations to counterfactuals, example-based explanations had no measurable impact on these evaluation metrics. Recent work from \citep{XUAN2025XAISubjective} supports these findings, stating that counterfactual explanations are perceived as less understandable than other types, such as feature importance, often seen as easier to grasp. However, explanations perceived as \enquote{easy to understand} were found to be both more intelligible and more misleading. This aligns with the findings of \citep{Chromik2021Loan}, suggesting that users might overestimate their understanding of local feature explanations due to the illusion of explanatory depth. Furthermore, these results are consistent with previous work \citep{Bucinca2020ProxyTasks,Wang2022AIConfidenceMultipleExplanationsDomainExpertise}, which demonstrates that subjective measures, such as user preferences, do not necessarily align or predict objective outcomes.
These results emphasize the importance of shifting from traditional feature-based explanations, which are commonly used in AI systems. Instead, we should adopt approaches that resemble human-like reasoning, such as counterfactuals. Furthermore, it is essential to integrate various types of explanations to offer complementary insights. This combination can address each explanation's shortcomings and limitations, ultimately leading to the development of personalized hybrid visualizations for explainable AI (XAI). 
Recent studies have proposed integrating actionable data-centric explanations \citep{Anik2021DataCentric,Liao2022HCXAI,Yurrita2023MergeExplanations,Esfahani2024AlgorithmicRecourse} alongside model-centric ones, offering potential benefits for both AI experts and lay users by connecting them to the training data and influencing their perceptions of trust and fairness in AI systems. For instance, research in the health domain has demonstrated that expert users gain significant advantages from hybrid explanations combining data-centric and global model-centric elements \citep{Bhattacharya2023DirectiveExplanations,Bhattacharya2024EXMOS,Bhattacharya2024ModelSteeringSystem,Szymanski2024HealthDashboardDataCentric}, though these approaches remain underexplored for lay users. Future work should focus on developing tailored explanation interfaces that adapt to users' expertise levels and contextual needs, ensuring both accessibility for lay users and depth for experts.
On top of this, tailoring XAI interfaces for users may involve assessing user-centric perspectives and characteristics, which we discuss in the next subsection.

\subsection{Individual Differences: NFC and Personalization in AI Interaction}
Our findings differ from previous work \citep{Millecamp2019NFC,Bucinca2021NFC,Conati2021NFC,BahelConati2024NFC}, which reported differences between low and high NFC individuals in terms of accuracy and cognitive load.
Interestingly, we found that both low and high NFC participants prioritized explanations (ranked 2nd) immediately after loan application attributes (ranked 1st), leaving AI information (ranked 3rd) as the least influential in decision-making. Moreover, low NFC individuals prioritized AI prediction over accuracy, while those with a high NFC seem to consider AI information as a whole.
We can identify two main reasons we might not have observed significant NFC-related differences compared to prior studies. 

First, the task's nature and complexity may have minimized the differences between NFC groups. Notably, prior studies focused on low-stakes tasks, such as explaining music recommendations \citep{Millecamp2019NFC}, nutrition choices in image-based domains \citep{Bucinca2021NFC}, and tutoring systems for university students with some domain knowledge \citep{BahelConati2024NFC}. In contrast, our study involved a high-stakes loan approval task using tabular data with eleven features, where participants were unfamiliar with the domain. Additionally, our explanations added substantial information for users to process, classifying the task as high-complexity according to \citep{Salimzadeh2023TaskComplexityDecisionMaking}. This suggests that as task complexity increases, NFC may lose its predictive ability to differentiate individual behaviors. 

Second, while the NFC personality trait has been shown to distinguish between low- and high-NFC individuals, it may not reliably explain differences in AI-driven decision outcomes, regardless of cognitive forcing. Recent AI-assisted user studies leveraging Large Language Models (LLMs) \citep{buçinca2024contrastiveexplanationsanticipatehuman} and Reinforcement Learning (RL) \citep{buçinca2024optimizinghumancentricobjectivesaiassisted} indicate that NFC may not always predict differences in users’ accuracy, learning, reliance on AI, or mental demand, regardless of explanation type or cognitive interventions. These findings highlight the need for alternative traits that might capture richer insights about intrinsic motivation to learn and think, such as Epistemic Curiosity \citep{Litman2008EC} or the five-dimensional curiosity scale \citep{Kashdan2018CEIII}.
Moreover, a notable methodological concern is dividing participants into low and high-trait groups after data collection based on the overall participant distribution median. This approach, commonly used for NFC and other traits, may lead to imbalances and unequal group sizes, complicating statistical analyses and consequent reproducibility of results. Future research should explore alternative user-centric metrics beyond personality traits that enable real-time categorization during studies, ensuring more balanced groups and dynamic personalization.

\subsection{Limitations and Future Work}
\label{sec:limitations}
We acknowledge three main limitations in our work. 
The first consists of using an AI model with uncalibrated confidence estimates. Although we assessed that calibration metrics did not improve the AI baseline model (Random Forest), this may have affected the computation of model confidence estimates and explanations generation, and consequently users' decision-making during the study. As such, we strongly encourage future studies to calibrate their AI models when necessary to ensure stability between AI probability outputs and confidence estimates. 
A second limitation is that our study employed a one-stage detection paradigm, where users' decision-making co-occurs with AI suggestions and explanations. 
While this approach mirrors many real-world applications applied to autonomous driving \citep{Atakishiyev24AutonomousDriving} and cybersecurity \citep{Desolda23phishing}, it may restrict the ability to disentangle users' independent reasoning from their reliance on AI advice. In contrast, two-stage detection paradigms, where users first evaluate a task independently before incorporating AI input, provide a clearer separation of cognitive engagement and reliance patterns. Future research should explore balancing these paradigms to achieve an optimal trade-off based on the target domain's specific demands, stakes, and cognitive complexity.
The last limitation concerns the generalizability of our findings beyond the specific domain, dataset, classification model, AI confidence split into low and high levels, and explanation methods used. Our study employed a publicly available loan approval dataset commonly used in research, along with a model achieving comparable evaluation metrics. 
While we ensured replicability by detailing the data processing, AI model, explanation generation, and statistical analysis, several variables unique to our setup may have influenced decision-making. Further, research is needed to evaluate the impact of AI and explanations across diverse domains with varying stakes and levels of uncertainty.

\section{Conclusion}
\label{sec:conclusion}
This article investigated how presenting AI information including prediction, confidence, accuracy, and explanation styles such as example-based, feature-based, rule-based, and counterfactual, affects users’ decision-making in loan approval tasks. Specifically, we conducted a user study ($N$ = 288) examining how these elements influence accuracy, reliance on AI, and cognitive load across six AI-assistance conditions: no AI, AI with no explanation, and AI with each of the four explanation styles. Additionally, given the recent interest in studying the Need for Cognition (NFC) personality trait in human-AI teams, we explored how NFC levels affect users’ prioritization of information, accuracy, and cognitive load when interacting with different explanation styles.

Our results show that high AI confidence significantly increases users' reliance on AI while reducing cognitive load, emphasizing the importance of accurately calibrating confidence estimates to reflect AI correctness. 
Counterfactual explanations, despite being rated as less understandable than other explanation types, overall increase users' accuracy, also reducing cognitive load and increasing reliance on AI, particularly when paired with correct AI predictions. In contrast, feature-based explanations failed to improve accuracy as anticipated. 
Moreover, we observed that NFC levels did not significantly differ in how users prioritize information or their reliance, accuracy, and cognitive load, suggesting that NFC's influence may be task- or context-specific.
These findings contribute to a deeper understanding of how AI-assisted decision-making can be optimized by integrating complementary explanation styles and tailoring interfaces to individual user needs. Future work should explore hybrid explanation systems and refine user-centric models with AI to create more adaptive, effective, and equitable human-AI collaboration frameworks.

% \section{Methods}
% \label{sec11}

% Topical subheadings are allowed. Authors must ensure that their Methods section includes adequate experimental and characterization data necessary for others in the field to reproduce their work. Authors are encouraged to include RIIDs where appropriate. 

% \textbf{Ethical approval declarations} (only required where applicable) Any article reporting experiment/s carried out on (i)~live vertebrate (or higher invertebrates), (ii)~humans or (iii)~human samples must include an unambiguous statement within the methods section that meets the following requirements: 

% \begin{enumerate}
% \item Approval: a statement which confirms that all experimental protocols were approved by a named institutional and/or licensing committee. Please identify the approving body in the methods section

% \item Accordance: a statement explicitly saying that the methods were carried out in accordance with the relevant guidelines and regulations

% \item Informed consent (for experiments involving humans or human tissue samples): include a statement confirming that informed consent was obtained from all participants and/or their legal guardian/s
% \end{enumerate}

% If your manuscript includes potentially identifying patient/participant information, or if it describes human transplantation research, or if it reports results of a clinical trial then  additional information will be required. Please visit (\url{https://www.nature.com/nature-research/editorial-policies}) for Nature Portfolio journals, (\url{https://www.springer.com/gp/authors-editors/journal-author/journal-author-helpdesk/publishing-ethics/14214}) for Springer Nature journals, or (\url{https://www.biomedcentral.com/getpublished/editorial-policies\#ethics+and+consent}) for BMC.

\bmhead{Acknowledgements}
This research is funded by the Italian Ministry of University and Research (MUR) and by the European Union - NextGenerationEU, Mission 4, Component 2, Investment 1.1, under grant PRIN 2022 PNRR "DAMOCLES: Detection And Mitigation Of Cyber attacks that exploit human vuLnerabilitiES” (Grant P2022FXP5B) — CUP: H53D23008140001. 

\section*{Declarations}

\noindent \textbf{Conflict of interest.} The authors declare no competing interests.

\begin{appendices}

\section{}
\label{sec:appendix}

\subsection{Model calibration}
\label{sec:model_calib}
Given we will show participants the RFC confidence for each prediction, we decided to calibrate the RFC probabilities before computing the confidence estimates using three methods: Isotonic Regression \citep{Zadrozny2001Isotonic}, Platt Scaling \citep{Platt2000Calib},  inductive and cross Venn-Abers \citep{Vovk2014VennAbers,Vovk2015PRoba,Manokhin2017VennAbers}. Specifically, we compared the RFC with ensembles of ten RFC models for each method to assess a ten-fold cross-validation. Nevertheless, in this specific scenario, these methods slightly worsened the metrics we took into consideration (Accuracy, Brier loss \citep{Brier1950VERIFICATIONOF}, Log loss \citep{Domingos1999LogLoss}, ROC-AUC \citep{Fawcett2004ROCAUC}, and Expected Calibration Error \citep{Chuan2017ECE}), except for the Isotonic Regression to some extent (see Table \ref{tab:model_calibration}).  We decided to use our original (uncalibrated) RFC model for the loan prediction task as it resulted in better calibration metrics than the other methods we used.

\begin{table}[!h]
    \centering
    \small
    \setlength{\tabcolsep}{5pt}
    \caption{Summary of the Random Forest calibration results using the following metrics: accuracy, Brier loss, Log loss, ECE, and ROC AUC. We omitted the inductive Venn-Abers given the worst results overall compared to the other methods. }
    \begin{tabular}{cccccc}
        \toprule
        \textbf{Method} & \textbf{Accuracy} & \textbf{Brier loss} & \textbf{Log loss} &
        \textbf{ECE} &
        \textbf{ROC AUC}\\
        \midrule
        RF raw probabilities & \textbf{0.8293} &	\textbf{0.1370}	& \textbf{0.4424} &	\textbf{0.0580}	& 0.8204\\ 
        Isotonic Regression & 0.8130 &	0.1403 &	0.4518	& 0.0618 &	\textbf{0.8215} \\  
        Platt Scaling & 0.8130 &	0.1413 &	0.4524 &	0.0768 &	0.8167 \\
        Cross Venn-Abers
       & 0.8211 &	0.1492	&0.4727 &	0.0641 &	0.8 \\
        \bottomrule
    \end{tabular}
    \label{tab:model_calibration}
\end{table}

\subsection{Need for Cognition Scale}
\label{sec:nfcs}
We will measure participants' Need for Cognition (NFC) with the NCS-6 considering a 5-point scale (1 = extremely uncharacteristic of
me; 5 = extremely characteristic of me). We will sum up all the six item scores and then compute the \textit{median} to split participants into low and high NFC.
We used the following six items to compute the NFC from \citep{LinsDeHolandaCoelho2020NFC6}\footnote{note: (R) = reversed items}:

\begin{enumerate}
    \item I would prefer complex to simple problems.
    \item I like to have the responsibility of handling a situation that requires a lot of thinking.
    \item Thinking is not my idea of fun. (R)
    \item I would rather do something that requires little thought than something that is sure to challenge my thinking abilities. (R)
    \item I really enjoy a task that involves coming up with new solutions to problems. 
    \item  I would prefer a task that is intellectual, difficult, and important to one that is somewhat important. 
\end{enumerate}

\subsection{Metrics Overview by Task} 
\label{sec:task_overview}

We summarized participants' performance on loan prediction tasks in Table \ref{sec:task_overview}, ordered by decreasing accuracy. Along with reliance on AI and cognitive load, we also reported participants' disagreement with correct AI advice, namely their under-reliance. We reported all the metrics in percent (\%), except for cognitive load.

\begin{table}[!h]
    \centering
    \footnotesize
    \setlength{\tabcolsep}{3.5pt}
    \caption{Participants' accuracy, reliance on AI, under-reliance on AI, and cognitive load for our loan prediction task instance settings.}

\begin{tabular}{cccccccc}
\toprule
\textbf{ID} &
\textbf{AI correctness} & \textbf{AI confidence} & \textbf{Accuracy} & \textbf{Reliance}
& \textbf{Under-reliance} & \textbf{Cognitive load}\\

\midrule
5 & correct  & high & 90.4 & 90.4 & 9.6 & 3.1 \\
1 & correct  & high & 85.4 & 85.4 & 14.6 & 3.3 \\
6 & correct  & low  & 71.2 & 71.2 & 28.7 & 3.8 \\
4 & correct  & low  & 56.2 & 56.2 & 43.8 & 3.7 \\
2 & correct  & low  & 44.2 & 44.2 & 55.8 & 3.8 \\
8 & wrong    & low  & 27.9 & 72.1 & - & 3.5 \\
3 & wrong    & high & 27.1 & 72.9 & - & 3.4 \\
7 & wrong    & high & 14.6 & 85.4 & - & 3.3 \\
\bottomrule
\end{tabular}
    \label{tab:tasks_overview}
\end{table}




\end{appendices}


\bibliography{sn-article}% common bib file

\end{document}